\pdfoutput=1

\documentclass[11pt]{article}

\usepackage[preprint]{acl}

\usepackage{times}
\usepackage{latexsym}

\usepackage[T1]{fontenc}

\usepackage[utf8]{inputenc}

\usepackage{microtype}

\usepackage{inconsolata}

\usepackage{graphicx}

\usepackage{amsmath,amsthm,amssymb,amsfonts}
\usepackage{subcaption}
\usepackage{float,stfloats}
\usepackage{rotating}
\usepackage{enumerate}
\usepackage{colortbl}
\usepackage{multirow}
\usepackage[framemethod=tikz]{mdframed}

\usepackage{hyperref}
\usepackage{mathtools}
\usepackage{makecell}
\usepackage{pifont}
\usepackage{tcolorbox}
\usepackage{bm}
\usepackage{booktabs}
\usepackage[normalem]{ulem}
\usepackage{enumitem}
\usepackage{array}
\newcommand*{\affaddr}[1]{#1}

\title{HAF-RM: A Hybrid Alignment Framework for Reward Model Training}

\author{{\normalsize Shujun Liu
$^\clubsuit$ ~ Xiaoyu Shen$^\spadesuit$ ~ Yuhang Lai$^\clubsuit$ ~Siyuan Wang$^\diamondsuit$~ Shengbin Yue$^{\clubsuit}$ \vspace{0.3mm} 
}\\ 
{\normalsize\bf Zengfeng Huang$^\clubsuit$ ~ Xuanjing Huang$^\clubsuit$ ~ Zhongyu Wei$^{\clubsuit}$\thanks{Corresponding author}\vspace{1mm}
}\\
\affaddr{\normalsize$^\clubsuit$Fudan University} \vspace{0.2mm}\\
\affaddr{\normalsize$^\spadesuit$Eastern Institute of Technology, Ningbo}\vspace{0.2mm}\\
\affaddr{\normalsize$^\diamondsuit$University of Southern California}\vspace{0.2mm}\\
\normalsize\texttt{\{shujunliu20,huangzf,xjhuang,zywei\}@fudan.edu.cn,}\\
\normalsize\texttt{xyshen@eitech.edu.cn,\{sbyue23,yhlai23\}@m.fudan.edu.cn}\\
\normalsize\texttt{
sw\_641@usc.edu}
}

\begin{document}
\maketitle
\begin{abstract}
The reward model has become increasingly important in alignment, assessment, and data construction for large language models (LLMs). Most existing researchers focus on enhancing reward models through data improvements, following the conventional training framework for reward models that directly optimizes the predicted rewards.
In this paper, we propose a hybrid alignment framework \textsc{HaF-RM} for reward model training by introducing an additional constraint on token-level policy probabilities in addition to the reward score. 
It can simultaneously supervise the internal preference model at the token level and optimize the mapping layer of the reward model at the sequence level.
Experiment results on five datasets sufficiently show the validity and effectiveness of our proposed hybrid framework for training a high-quality reward model.
By decoupling the reward modeling procedure and incorporating hybrid supervision, our \textsc{HaF-RM} framework offers a principled and effective approach to enhancing the performance and alignment of reward models, a critical component in the responsible development of powerful language models. We release our code at \href{https://haf-rm-anonymized.github.io}{https://haf-rm-anonymized.github.io}.
\end{abstract}
\newcommand{\p}[1]{\left({#1}\right)}
\newcommand{\B}[1]{\left[{#1}\right]}
\newcommand{\C}[1]{\left\{{#1}\right\}}
\newcommand{\haf}{\textsc{HaF} }
\newcommand{\pp}{\phantom{1}}
\section{Introduction}
\label{Introduction}
Recent periods have witnessed a continuous evolution of Large Language Model (LLM) techniques, especially in pre-training~\cite{bert,pretrain,lmarefewshotlearners} and instruction tuning~\cite{Wei2021FinetunedLM,selfInstruct,yue2023disc}. As these models advance, researchers have shifted their focus from generating correct responses to aligning outputs more closely with human preferences~\citep{valuealign} through Reinforcement Learning from Human Feedback (RLHF)~\citep{ouyang2022training}. 
As an efficient alternative to human feedback, reward models for generative language models emerge, facilitating scalable alignment in training~\citep{DeepRLhumanf,summarizefromHF}, response generation~\citep{scalinglawforrm,controlledgene,regularizedbon}, and data construction\citep{RRHF} etc.

\begin{figure}[t]
\centering
\includegraphics[width=0.45\textwidth]{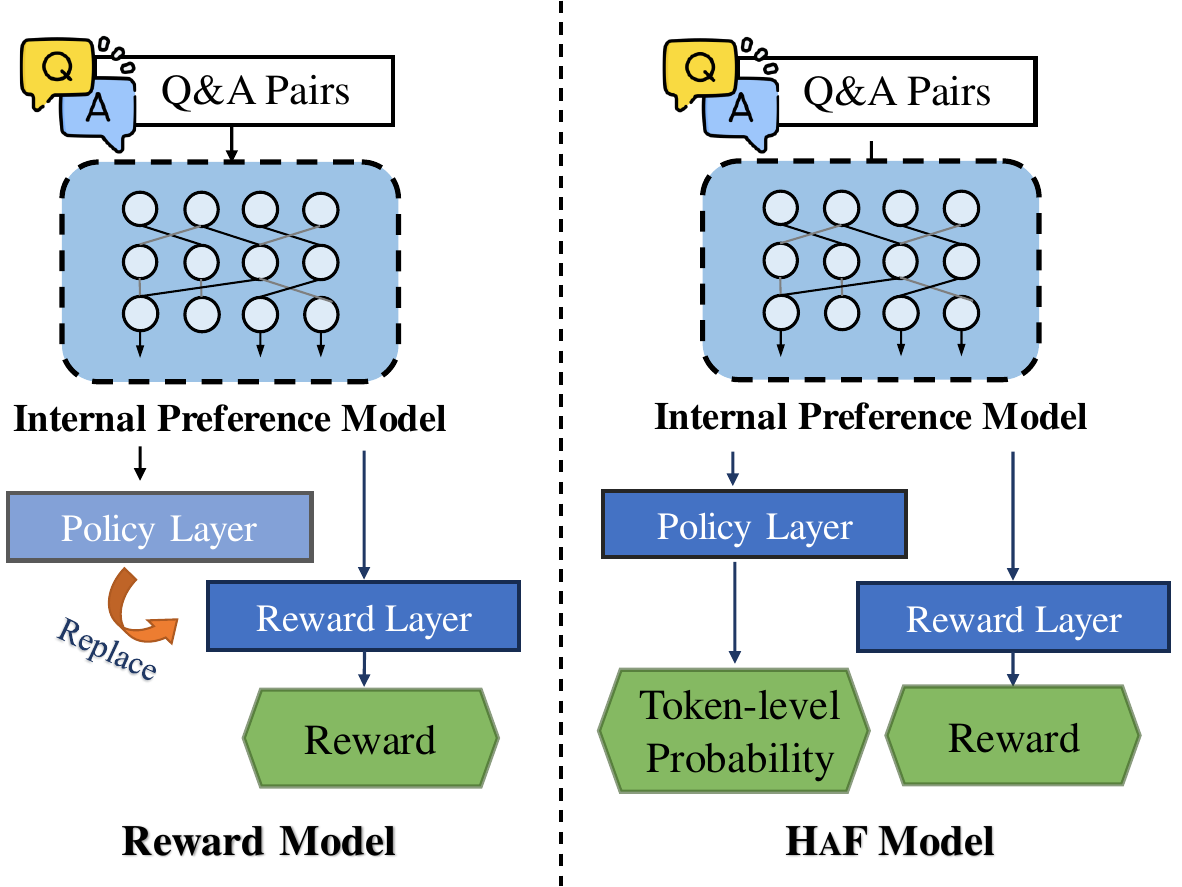}
\caption{\haf model structure. It retains the policy layer which outputs the token-level probability.}
\label{fig-model structure}
\end{figure}

Despite the availability of numerous sophisticated reward models~\citep{OpenAssistant,starling2023}, several key limitations remain. 
First, many reward models are proprietary and closed-source, originating from industry, which restricts their further training and transfer. Second, prior studies have highlighted incorrect and ambiguous preferences within the training data of these reward models~\citep{TrainingAH,failurelearningRM}. These two issues both limit the quality and generalizability of existing reward models, necessitating further enhancement either from the data perspective or the training process. Recent efforts primarily focus on enriching data sources to improve reward models, including incorporating external tools or information sources to enhance generalization~\citep{ToolAugmentedRM,factuallyaugmented} or leveraging fine-grained signals~\citep{finegrained,drlc} and their combinations~\citep{compositional,alarm}. In contrast, this work aims to improve the training framework of reward models.

A reward model is typically structured with two components: a transformer-based model (referred to as the ``internal preference model''), and a projection module called ``reward layer''~(usually a linear layer). The former outputs preference vectors for each token, while the latter maps these vectors to sequence-level rewards.
We argue that the standard practice for training the reward model may cause insufficient supervision for preference modeling, which can be improved by performing hybrid supervision of both token-level and sequence-level.

Given that a policy model also relies on an internal preference model to predict expected rewards for each action or token, essentially acting as a \textit{Q}-function under token-level supervision~\cite{DPOQFunc}, we propose a Hybrid Alignment Framework (HAF). This framework jointly optimizes the reward model and the policy model by sharing the internal preference model. With an additional policy loss, we can directly supervise the internal preference model at the token level, while simultaneously optimizing the mapping layer of the reward model using the reward loss, enabling more effective alignment of the reward model. 

We provide massive empirical experiments with an intuitional justification to demonstrate the effectiveness of our HAF. 
In the experiment section, we compare the performance of reward models trained using our framework against those resulting from traditional baseline and DPO approaches across five public datasets. The results highlight the advantage of HAF with different policy losses integrated. Further analysis reveals that using additional policy loss can improve the performance of policy model calibration, which opens a new horizon for training high-quality reward models.

\section{Hybrid Alignment Framework}\label{sec:theoretical analysis}
In this section, we first introduce the necessary notations~(Section~\ref{sec:notation}). Then we derive the formation of reward loss and policy loss as well as their practical calculation methods~(Section~\ref{sec:loss function}), and propose \haf to effectively utilize the similarity between the reward model and the policy model~(Section~\ref{sec:haf implementation}). Finally, we provide an intuition-based explanation for why \haf works~(Section\ref{sec:analysis}).
\subsection{Notation}
\label{sec:notation}
The objective of our framework is to train the reward model~$\bm{r}$ based on a pairwise comparison dataset~(also known as ``preference dataset'')~$\mathcal D$, following typical reward model training settings. 
\begin{itemize}[leftmargin=12pt,itemsep=2pt]
    \item $\mathcal{D}=\C{\p{x_i,y_i,y_i'}}_{i=1}^n$ represents the dataset used to train the reward model, where $x_i$, $y_i$ and $y_i'$ are the query, preferred and non-preferred responses respectively.
    \item $\mathcal P=\C{\p{x,y}\mid\p{x,y,y'}\in\mathcal D}\cup\left\{\p{x,y'}\mid\right.\\\left.\p{x,y,y'}\in\mathcal D\right\}$ is the set of query-response pairs from the dataset $\mathcal D$. 
    \item $\bm{r}$ is the \textbf{reward model} which can be split into two parts as $\bm{r}\mathrm{\p{x,y}}=\mathrm{F}\circ\phi\p{x,y}$, to output the reward of a response $y$ given a query $x$. Here, $\phi\p{\cdot,\cdot}$ denotes the model's internal preference model, while $\mathrm{F}$ serves as the reward prediction layer mapping the model's internal preference to the final reward. We use the symbol $\circ$ to signify function nesting, i.e., $\mathrm{F}\circ \phi\p{x,y}=\mathrm{F}\p{\phi\p{x,y}}$.
    \item $\bm{\pi}$ is the \textbf{policy model}, and $\bm{\pi}\p{x,y}$ is the generation probability of $y$ given $x$. It can also be divided into two parts as $\bm{\pi}\p{x,y}=\mathrm{K}\circ\phi\p{x,y}$ where the policy prediction layer $\mathrm{K}$ maps the model's internal preference to the generation probability. 
    \item The \textbf{Oracle value} is denoted as the corresponding letter with an asterisk such as $\bm{r}^*$(Oracle reward model), $\phi^*$(Oracle model preference), $\mathrm{F}^*$(Oracle reward prediction layer) and $\mathrm{K}^*$(Oracle policy prediction layer). 
\end{itemize}

\subsection{Basic Loss Functions}
\label{sec:loss function}
We use $\mathrm D_1$ to represent the distribution discrepancy between the reward model's output and the oracle reward model's output, and $\mathrm D_2$ for the outputs of the policy model and the oracle policy model.

\paragraph{Reward Loss} 
The standard reward loss~$\mathcal L_s$ considers the precision of rewards alone, being a simple and direct metric to quantify the quality of a reward model. 

\begin{equation}
\mathcal{L}_s\coloneqq\underset{d}{\mathbb{E}}\B{\mathrm{D}_1\p{\bm{r}\p{d},\bm{r}^*\p{d}}}
\label{eq:def reward loss}    
\end{equation}
We use $d$ to denote $\p{x,y}$ for notational simplicity.

In avoiding the issue of uncertain reward values, there is consensus on the use of the Bradley-Terry model~\citep{BTmodel} to transform the reward modeling problem into a probability optimization problem~\citep{summarizefromHF,DPO,simpo}, which yields the popular form of a binary classification cross-entropy loss:
\begin{equation}
\begin{aligned}
\mathcal{L}_s\leftarrow\underset{\p{x,y,y'}\sim\mathcal D}{\mathbb E}&\left[-\log\sigma\p{\bm{r}\p{x,y}-\bm{r}\p{x,y'}}\right]
\end{aligned}
\label{eq:standard_reward}
\end{equation}
where $\sigma\p{\cdot}$ is the sigmoid function~(derivation can be found in Appendix~\ref{app:reward loss}).

\paragraph{Policy Loss}
Similar to the reward loss, the standard policy loss aims to measure the error of the policy model.

\begin{equation}
\mathcal{L}_P\coloneqq\underset{d}{\mathbb{E}}\B{\mathrm{D}_2\p{\bm{\pi}\p{d},\bm{\pi}^*\p{d}}}
\label{eq:def policy loss}    
\end{equation}

Here, we use DPO~\cite{DPO} for calculating policy loss since its derivation is similar to that made for the reward loss~(as detailed in Appendix~\ref{app:dpo as policy loss}).
\begin{equation}
\begin{aligned}
\mathcal{L}_P\leftarrow\underset{\p{x,y,y'}\sim\mathcal D}{\mathbb E}&\B{-\log\sigma\p{\tau\p{pd_{win}-pd_{lose}}}}
\end{aligned}
\label{eq:dpo_as_policy_loss}
\end{equation}
$\footnotesize pd_{win}=\log\frac{\bm{\pi}(x,y)}{\bm{\pi}_{ref}(x,y)}$, $\footnotesize pd_{lose}=\log\frac{\bm{\pi}(x,y')}{\bm{\pi}_{ref}(x,y')}$. $\bm{\pi}_{ref}$ is the reference policy model and $\tau$ is the hyperparameter set to 0.1. 

\begin{figure*}[t]
\centering
\includegraphics[width=0.95\textwidth]{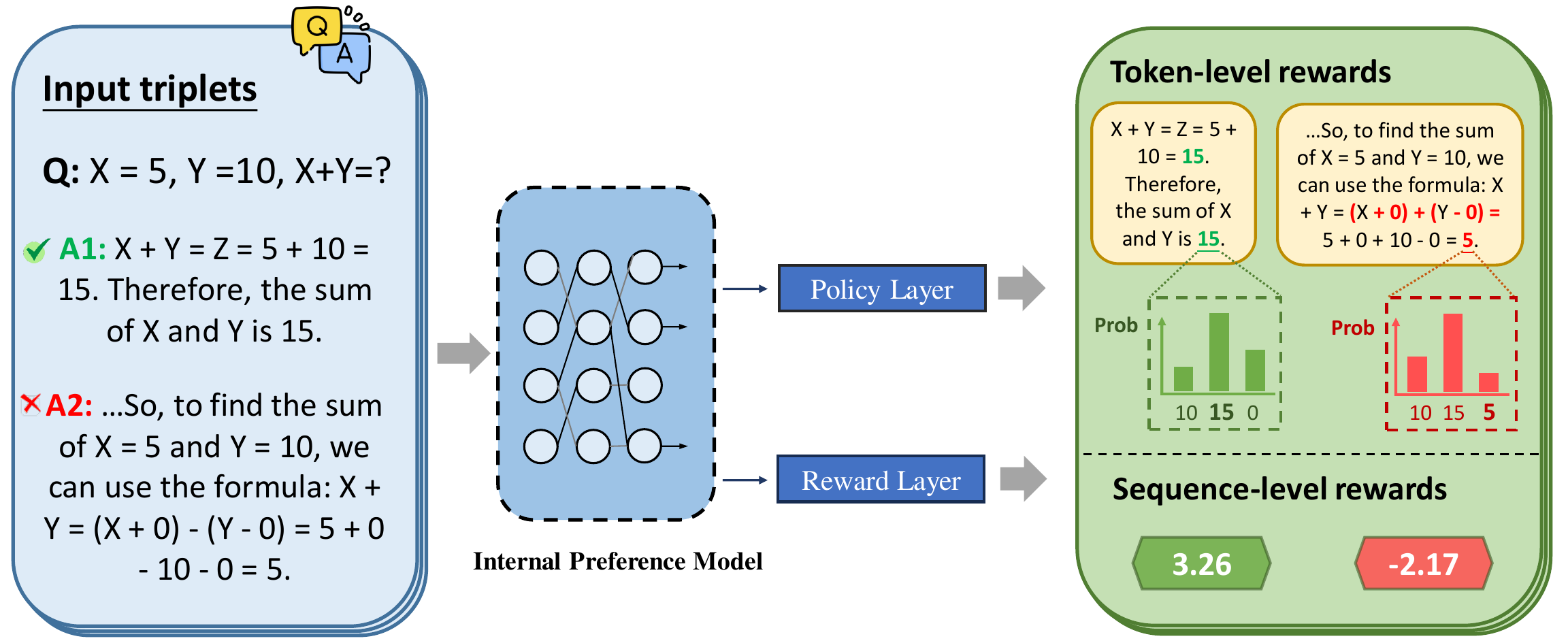}
\caption{\haf training framework. We add the reward layer to the language model while retaining its policy layer. During training, we optimize both the token-level rewards and sequence-level rewards for the input triplets by maximizing the reward differences between better responses and worse responses.}
\label{intro pic}
\end{figure*}

\begin{figure}[t]
\centering
\includegraphics[width=0.48\textwidth]{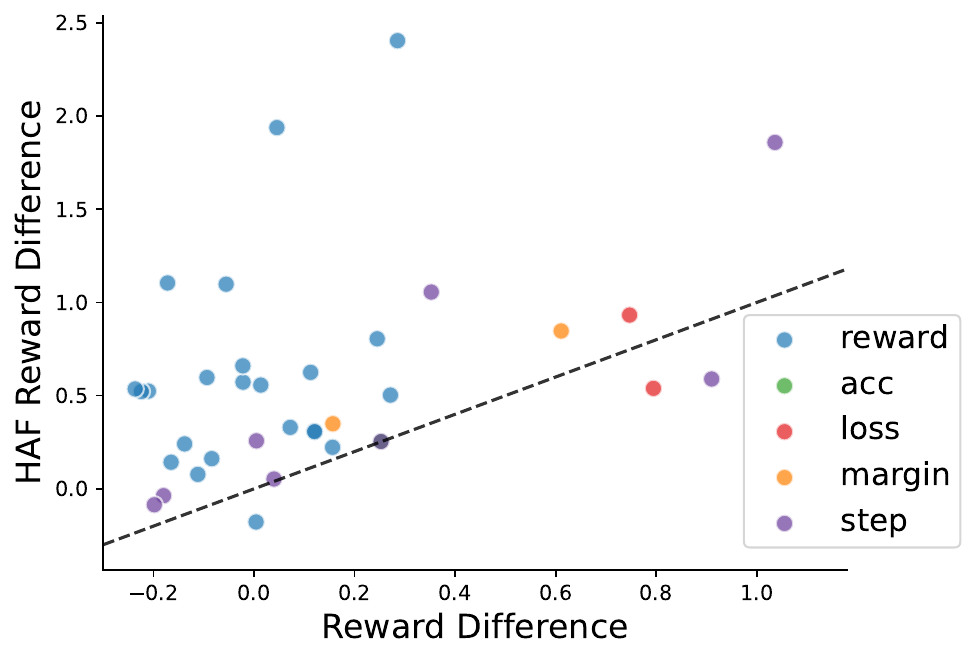}
\caption{\haf tends to assign higher scores to the responses it generates. The x-axis represents the score difference between the ideal reward model's evaluation of the content generated by \textsc{HaF}'s policy head and the content generated by the model trained with DPO. The y-axis indicates the score difference when \haf evaluates these two outputs. Different colors represent different model checkpoint selection strategies.}
\label{fig:consistency}
\end{figure}

\subsection{\haf Implementation}
\label{sec:haf implementation}
\paragraph{Hybrid Alignment Loss}
To fully leverage the similarity between the reward model and the policy model, we incorporate an additional supervising term $\mathrm D_2$ on the policy model into the loss function.
By calibrating the shared preference space, we effectively align the model in a hybrid manner:
\begin{equation}
\begin{aligned}
\mathcal{L}_H\coloneqq\underset{d}{\mathbb E}\left[\mathrm D_1\p{\bm{r}\p{d},\bm{r}^*\p{d}}\right.\qquad\qquad\,\;&\\
+\alpha\cdot\left.\mathrm D_2\p{\bm{\pi}\p{d},\bm{\pi}^*\p{d}}\right]\qquad\quad\;\;&\\
=\underset{d}{\mathbb E}\left[\mathrm D_1\p{\mathrm{F\circ\phi}\p{d},\mathrm{F^*\circ\phi^*}\p{d}}\right.\,\;\;&\\
+\alpha\cdot\left.\mathrm D_2\p{\mathrm{K\circ\phi}\p{d},\mathrm{K^*\circ\phi^*}\p{d}}\right]&
\end{aligned}
\label{eq:calibrated_reward}
\end{equation}
where $\alpha$ is a hyperparameter to balance losses from the reward and policy model, $\phi$ is the shared internal preference model which receives gradients from both loss terms. 

\paragraph{Model structure}
The most commonly used decoder-only LLM consists of stacked transformer blocks~\citep{attentionisall} or similar structures, and a linear layer 
for policy projection. 
In the reward model, only the shape of the final linear layer is adjusted to match the format of the reward value output compared to the policy model~\citep{summarizefromHF}. We retain two linear layers for our model, enabling it to output rewards and probabilities simultaneously, as shown in Figure~\ref{fig-model structure}. 

\subsection{Why \haf is Better?}
\label{sec:analysis}

Figure~\ref{fig:consistency} shows the consistency between the reward model and the policy model in preference learning. Despite possessing similar generation quality, the policy model which shares parameters with the reward model is rated higher, indicating that the two models do have resembling preferences when they have the same internal preference model. We will elaborate on this finding in Appendix~\ref{app:consistency}.

Besides, we provide an intuitive explanation of why the hybrid alignment loss can yield a better solution than simply using the standard reward loss.

\paragraph{Claim 1.}
\emph{The model learned from the joint calibrated loss outperforms the one learned solely from the preference space using the standard reward loss}. Details can be found in Appendix~\ref{app:math enlight}.

\paragraph{Claim 2.}
\emph{Policy loss can act as a regularization term preventing the inner representation from degrading, so HAF tends to outperform the traditional training framework.}

\section{Experimental Setup}

\subsection{Datasets}

We comprehensively evaluate the performance of our framework using five public datasets: Anthropic-HH-Harmless (HH-harmless)~\citep{TrainingAH},  Anthropic-HH-Helpful (HH-Helpful)~\citep{TrainingAH}, Beaver Safe (BS)~\citep{Ji2023BeaverTailsTI}, Alpaca Human Pref (AHP) \citep{AlpacaFarm}, and Chatbot Arena (CA)~\citep{LLMasaJudge}. Since AHP and CA do not provide original data split for evaluation, we randomly extract 10\% from the original data as the test set. Detailed statistics of our used datasets for training are shown in Table \ref{tab:ds statistic}. 
\begin{table}[!h]
\centering
\resizebox{0.47\textwidth}{!}{
\begin{tabular}{@{}clccc@{}}
\toprule
&\multicolumn{1}{c}{Dataset}    & Size & \#Word/QA & \#Token/QA \\ \midrule
&Harmless       & 12,915              & 42.9                   & 61.5                    \\
&Helpful        & 13,543              & 54.3                   & 77.2                    \\
&BS      & 47,625              & 69.3                   & 88.5                    \\
&AHP & \phantom{0}8,722               & 59.6                   & 81.9                    \\
&CA     & 19,466              & 165.5\phantom{0}                  & 257.6\phantom{0}                   \\ \bottomrule
\end{tabular}
}
\caption{Statistics of the Training Subsets.}
\label{tab:ds statistic}
\end{table}



\subsection{Compared Models}
\label{Implementation Details}

\paragraph{Baseline}
We compare our framework with the standard training approach, wherein the reward model only has a reward layer dedicated to reward prediction and is optimized only with reward loss, as delineated in Eq.~\ref{eq:standard_reward}.

\paragraph{DPO}
DPO can implicitly convert model's outputs into reward values~\citep{DPO}, so the model can also function as a reward model~\citep{DPOQFunc}. Following the work of \citet{rewardbench}, we evaluate the model trained with DPO loss.

\paragraph{HAF}
Under our framework, the reward model has both a reward layer and a policy layer for predicting sequence-level rewards and providing token-level probabilities. 

Our framework is implemented based on three different backbone LLMs including both pre-trained and fine-tuned models: \texttt{Phi-2-2.7B}~\cite{phi2}, \texttt{Mistral-7B-base-v0.3} and \texttt{Mistral-7B-Instruct-v0.2}~\cite{mistral7b}. We train Phi-2 and Mistrals using full-parameter and Low-rank Adaptation (LoRA) \cite{lora} strategies, respectively. 
More implementation details can be found in Appendix~\ref{app:setup}.

\section{Experiment Results}
\label{app:Results}
\subsection{Intrinsic Performance of Reward Models}

The primary function of a reward model is to evaluate the quality of responses to a given question, which involves accurately comparing pairs of answers to the same question. To demonstrate the effectiveness of our \haf in training reward models, we first conduct several experiments evaluating the intrinsic performance of our trained reward model, specifically by taking judgment accuracy as the evaluation metric. 

\subsubsection{Overall Performance}

\begin{table*}[t]
\centering
\resizebox{0.7\textwidth}{!}{
\begin{tabular}{@{}clccccccc@{}}
\toprule
&\multicolumn{1}{c}{Method}& \makecell{Helpful} & \makecell{Harmless} & \makecell{CA} & \makecell{BS} & \makecell{AHP} & \textbf{Avg}&  \\ 
\midrule
&DPO(Phi-2) & \underline{69.70} & 66.30 & 66.80 & \textbf{87.80} & 52.60 & 68.64&\\
&Baseline(Phi-2) & 64.30 & \underline{69.50} & \textbf{79.30} & 76.00 & \underline{58.40} & 69.50 & \\
&\haf(Phi-2) & \textbf{76.40} & \textbf{70.40} & \underline{79.00}  & \underline{84.00} & \textbf{60.80} & \textbf{74.12} &\\
\midrule
&DPO(Mistral-base) & 64.60 & \underline{69.90} & \underline{68.80} & \textbf{91.70} & \underline{53.80} & 69.76 &\\
&Baseline(Mistral-base) & \underline{72.60} & 69.80 & 64.20 & 78.30 & 50.40 & 67.06 &\\
&\haf(Mistral-base) & \textbf{73.00} & \textbf{70.00} & \textbf{74.40} & \underline{85.40} & \textbf{56.30} & \textbf{71.82} &\\
\midrule
&DPO(Mistral) & 74.29 & 70.30 &  \textbf{81.90} & \textbf{92.70}& \underline{60.30} & 75.90 &\\
&Baseline(Mistral) & \textbf{76.20} & \underline{72.70} & 79.80 & 80.80 & 56.30 & 73.16 &\\
&\haf(Mistral) & \underline{75.80} & \textbf{73.10} & \textbf{81.90} & \underline{88.70} & \textbf{63.10} & \textbf{76.52} &\\
 \bottomrule
\end{tabular}
}
\caption{Overall results (accuracy) for each dataset, by calculating the proportion that the better response is scored higher. The best performance is highlighted in boldface and the suboptimal result is underlined.
}
\label{tab:overall accres}
\end{table*}


Table~\ref{tab:overall accres} presents the overall results of our \haf compared to two basic approaches across five datasets.
We observe that \textbf{DPO and the baseline method show similar performance on average but there is significant variability in individual comparisons}. This suggests that the two methods focus on different features when learning preferences. In contrast, HAF consistently outperforms both, indicating its ability to effectively integrate features from both approaches to better learn preferences.

Specifically, Mistral-base performs poorly on the Helpful, CA, and AHP datasets because these datasets require preferences related to the quality of responses. \textbf{Since the base model has not undergone instruction tuning, it lacks the representation of relevant features, making it difficult to accurately judge response quality}. In contrast, the extensively trained base model is capable of distinguishing between benign and harmful content, allowing it to perform comparably to Mistral-Instruct on the safety-related BS and Harmless datasets. Nevertheless, \haf demonstrates promising results even for these challenging preferences.


Notably, DPO achieves the highest performance on BS across all three models, which is probably caused by DPO's ``concentrated'' data-fitting manner~\citep{IPO}. This is evident from the much lower variance in token-level perplexity for good and bad responses in the BS dataset compared to other datasets, indicating a more concentrated distribution respectively of these two subsets~(refer to Appendix~\ref{app:overallperformance} for detailed illustration). By integrating DPO loss, our \haf partially captures this ``concentrated'' data-fitting characteristics, leading to a more nuanced improvement on BS compared to the baseline methods. However, DPO's concentrated data-fitting may potential lead to over-fitting issues, whereas \haf and the baseline demonstrate better generalization ability, which we will elaborate on in the following experiments.


 
\begin{figure*}[t]
\centering
\includegraphics[width=1\textwidth]{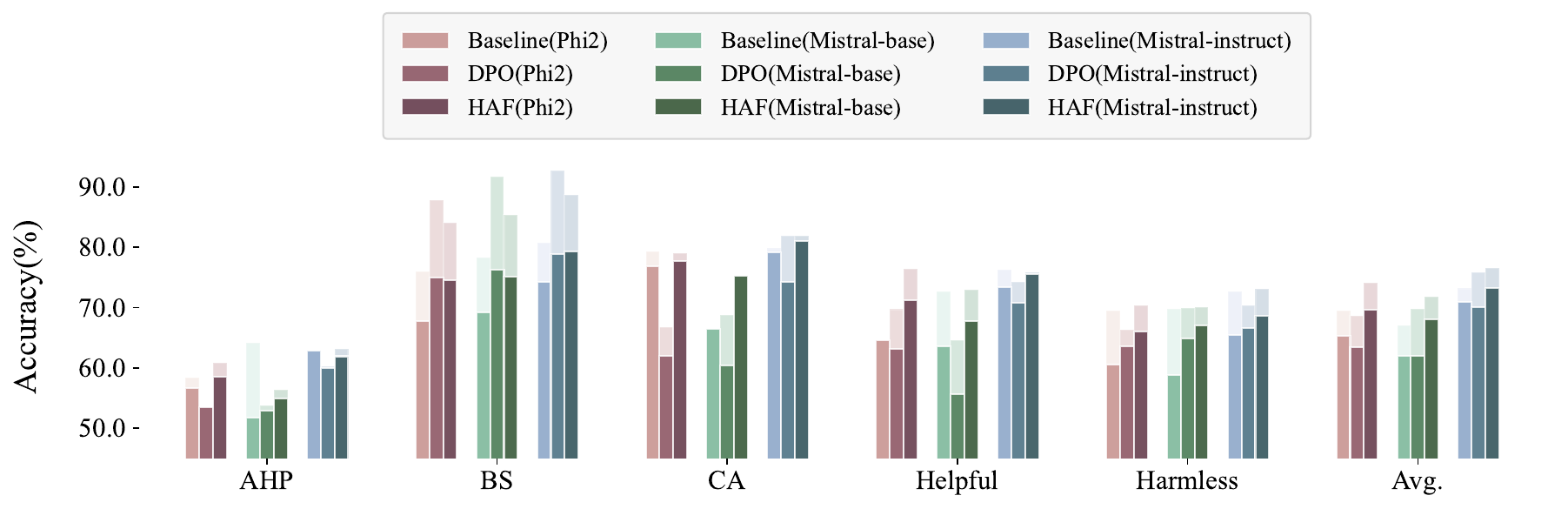}
\caption{The performance differences of \haf/ baseline / DPO under mixed preference training, with light shading indicating the upper bound performance of individually trained reward models on each dataset.}
\label{fig:mixed_data}
\end{figure*}

\begin{table*}[!ht]
\centering

\renewcommand\arraystretch{1.7}
\resizebox{0.96\textwidth}{!}{
\resizebox{\linewidth}{!}
{
\begin{tabular}{@{}ccccccc@{}}
\toprule
Acc(\%) & AHP$_C$        & CA$_C$          & Helpful$_C$         & BS$_S$    & Harmless$_S$ & Avg. \\[0.9ex]
\hline
\multicolumn{7}{c}{\cellcolor[HTML]{DEDCDC}\textit{internal}}\\ 
Phi-2
&$\textrm{67.50}^{_{(\textrm{1.20}\uparrow)}}_{^{(\textrm{23.70}\uparrow)}}$
&$\textrm{62.45}^{_{(\textrm{1.35}\downarrow)}}_{^{(\textrm{11.60}\uparrow)}}$  &$\textrm{66.10}^{_{(\textrm{0.90}\uparrow)}}_{^{(\textrm{19.80}\uparrow)}}$
&$\textrm{70.60}^{_{(\textrm{5.60}\uparrow)}}_{^{(\textrm{4.60}\uparrow)}}$  &$\textrm{76.90}^{_{(\textrm{1.50}\uparrow)}}_{^{(\textrm{8.60}\uparrow)}}$
&$\textrm{68.71}^{_{(\textrm{1.57}\uparrow)}}_{^{(\textrm{13.66}\uparrow)}}$\\
\hline
Mistral-base
&$\textrm{59.65}^{_{(\textrm{4.90}\uparrow)}}_{^{(\textrm{14.55}\uparrow)}}$
&$\textrm{56.35}^{_{(\textrm{2.15}\downarrow)}}_{^{(\textrm{6.00}\uparrow)}}$
&$\textrm{62.40}^{_{(\textrm{0.85}\downarrow)}}_{^{(\textrm{12.85}\uparrow)}}$
&$\textrm{69.60}^{_{(\textrm{0.50}\uparrow)}}_{^{(\textrm{3.30}\uparrow)}}$
&$\textrm{75.30}^{_{(\textrm{1.90}\uparrow)}}_{^{(\textrm{5.80}\uparrow)}}$
&$\textrm{64.66}^{_{(\textrm{0.86}\uparrow)}}_{^{(\textrm{8.50}\uparrow)}}$\\\hline
Mistral
&$\textrm{72.20}^{_{(\textrm{8.40}\uparrow)}}_{^{(\textrm{12.75}\uparrow)}}$  &$\textrm{63.30}^{_{(\textrm{0.70}\downarrow)}}_{^{(\textrm{9.65}\uparrow)}}$  &$\textrm{67.40}^{_{(\textrm{0.20}\downarrow)}}_{^{(\textrm{14.25}\uparrow)}}$  &$\textrm{71.90}^{_{(\textrm{1.40}\uparrow)}}_{^{(\textrm{3.00}\uparrow)}}$  &$\textrm{76.70}^{_{(\textrm{2.40}\uparrow)}}_{^{(\textrm{5.70}\uparrow)}}$
&$\textrm{70.30}^{_{(\textrm{2.26}\uparrow)}}_{^{(\textrm{9.07}\uparrow)}}$\\
 \hline
\multicolumn{7}{c}{\cellcolor[HTML]{DEDCDC}\textit{external}}                                      \\ 
Phi-2
&$\textrm{85.14}^{_{(\textrm{1.36}\uparrow)}}_{^{(\textrm{65.88}\uparrow)}}$
&$\textrm{95.27}^{_{(\textrm{0.34}\uparrow)}}_{^{(\textrm{19.59}\uparrow)}}$  &$\textrm{89.86}^{_{(\textrm{6.08}\uparrow)}}_{^{(\textrm{74.66}\uparrow)}}$
&$\textrm{66.30}^{_{(\textrm{0.95}\uparrow)}}_{^{(\textrm{2.04}\uparrow)}}$  &$\textrm{66.44}^{_{(\textrm{0.38}\downarrow)}}_{^{(\textrm{4.62}\uparrow)}}$
&$\textrm{80.60}^{_{(\textrm{8.35}\uparrow)}}_{^{(\textrm{33.36}\uparrow)}}$\\
\hline
Mistral-base
&$\textrm{79.66}^{_{(\textrm{20.34}\uparrow)}}_{^{(\textrm{64.14}\uparrow)}}$
&$\textrm{93.79}^{_{(\textrm{21.03}\uparrow)}}_{^{(\textrm{33.45}\uparrow)}}$
&$\textrm{81.38}^{_{(\textrm{6.90}\downarrow)}}_{^{(\textrm{67.24}\uparrow)}}$
&$\textrm{70.40}^{_{(\textrm{3.27}\uparrow)}}_{^{(\textrm{8.73}\uparrow)}}$
&$\textrm{63.30}^{_{(\textrm{3.82}\downarrow)}}_{^{(\textrm{4.91}\uparrow)}}$
&$\textrm{77.70}^{_{(\textrm{6.79}\uparrow)}}_{^{(\textrm{35.69}\uparrow)}}$\\\hline
Mistral
&$\textrm{91.55}^{_{(\textrm{32.77}\uparrow)}}_{^{(\textrm{53.37}\uparrow)}}$
&$\textrm{91.89}^{_{(\textrm{3.04}\uparrow)}}_{^{(\textrm{16.21}\uparrow)}}$  &$\textrm{82.43}^{_{(\textrm{1.69}\uparrow)}}_{^{(\textrm{63.51}\uparrow)}}$
&$\textrm{70.52}^{_{(\textrm{1.22}\downarrow)}}_{^{(\textrm{4.08}\uparrow)}}$  &$\textrm{73.37}^{_{(\textrm{2.72}\uparrow)}}_{^{(\textrm{5.17}\uparrow)}}$
&$\textrm{81.95}^{_{(\textrm{7.80}\uparrow)}}_{^{(\textrm{28.47}\uparrow)}}$\\
\bottomrule
\end{tabular}
}
}
\caption{Results for out-of-distribution data. Subscripts $C$ and $S$ denote the subjects of training sets, where $C$ represents Chat and $S$ represents Safety. ``\textit{internal}'' refers to testing results among datasets sharing the same subject category, while ``\textit{external}'' refers to testing results on RewardBench. The displayed accuracies are for \haf, with superscripts and subscripts indicating the performance differences relative to the baseline and DPO, respectively. $\uparrow$ denotes an improvement with \haf, while $\downarrow$ signifies a decline.}
\label{tab:OOD study}
\end{table*}

\subsubsection{Evaluation on Mixed Data}
To illustrate \haf's effectiveness in training reward models on mixed data, we construct a dataset by evenly sampling and combining examples from all five datasets. As shown in Figure \ref{fig:mixed_data}, our proposed hybrid alignment framework achieves the best overall performance across all reward models when evaluated on the mixed data distribution. This suggests that \haf is more effective at learning the diversity within the combined datasets.

Specifically, compared to the individual results on corresponding datasets in Table~\ref{tab:overall accres}~(shown as lightly shaded bars in Figure~\ref{fig:mixed_data}), we observe that \textbf{both the baseline method and \haf replicate their performance in learning individual preferences better than DPO when applied to mixed preference learning}. Notably, DPO's performance drops significantly on the CA and Helpful datasets, suggesting that DPO tends to fit the most prominent features of the overall data distribution. This also aligns with the finding of \citet{chen2024preference} that DPO would optimize the margins of correct data rather than the wrong ones.

\subsubsection{Transferability to OOD Data}
We further evaluate the generalizability of our framework to entirely held-out out-of-distribution (OOD) datasets to simulate distribution shifts in real-world applications.
Specifically, the five datasets are grouped into two categories: ``Safety''~(BS, Harmless) and ``Chat''~(AHP, CA, Helpful). We train the model on one dataset and evaluate its performance within the same category. The evaluation data comes from two sources, including the ``\textit{internal}'' source referring to different datasets within the same category, and an ``\textit{external}'' source, consisting of test data on related topics from RewardBench. 


As shown in Table~\ref{tab:OOD study}, \haf achieves a higher \textit{internal} accuracy compared to both Baseline and DPO, demonstrating \haf’s strong ability to learn preferences and effectively generalize to similar preference distributions, even with notable differences in language style and topic. As \citet{llama2} noted, RLHF causes distributional shifts in the policy model during training, often requiring iterative training of the reward model. \haf’s robustness against these distributional shifts could potentially be a key factor in mitigating this issue. 

It is important to note that nearly all of DPO’s test outcomes converge around 50\%, indicating a complete loss of modeling capability for OOD data. This likely stems from DPO's inherent nature as a language model, where the generation process exhibits strong stylistic biases, favoring responses that align with its style (as reflected in generation probabilities and implicit reward values). When response distribution deviates from these stylistic norms (e.g., responses that are too short, too long, or use different vocabulary), DPO's output probabilities become highly inaccurate, rendering it unsuitable as a conventional reward model.

From these three experiments, we conclude that DPO learns features significantly different from those learned by the baseline method. In contrast, \haf inherits both the baseline method's generalization ability and DPO's stronger fitting capability.

\subsection{Extrinsic Evaluation on Downstream Task}
Intrinsic performance metrics offer only a partial view of a reward model's efficacy. To comprehensively assess their practical applicability in real-world scenarios, it is crucial to evaluate how these models perform in downstream tasks that closely simulate practical applications.

In this section, we evaluate the robustness and effectiveness of \haf in such scenarios. Specifically, we explore its performance in two distinct downstream tasks: best-of-N sampling, a training-free response generation strategy~\citep{summarizefromHF,scalinglawforrm,regularizedbon}, and RLHF, a training-dependent alignment method. 

\subsubsection{Best-of-N}

\begin{figure}[t]
\centering
\includegraphics[width=0.48\textwidth]{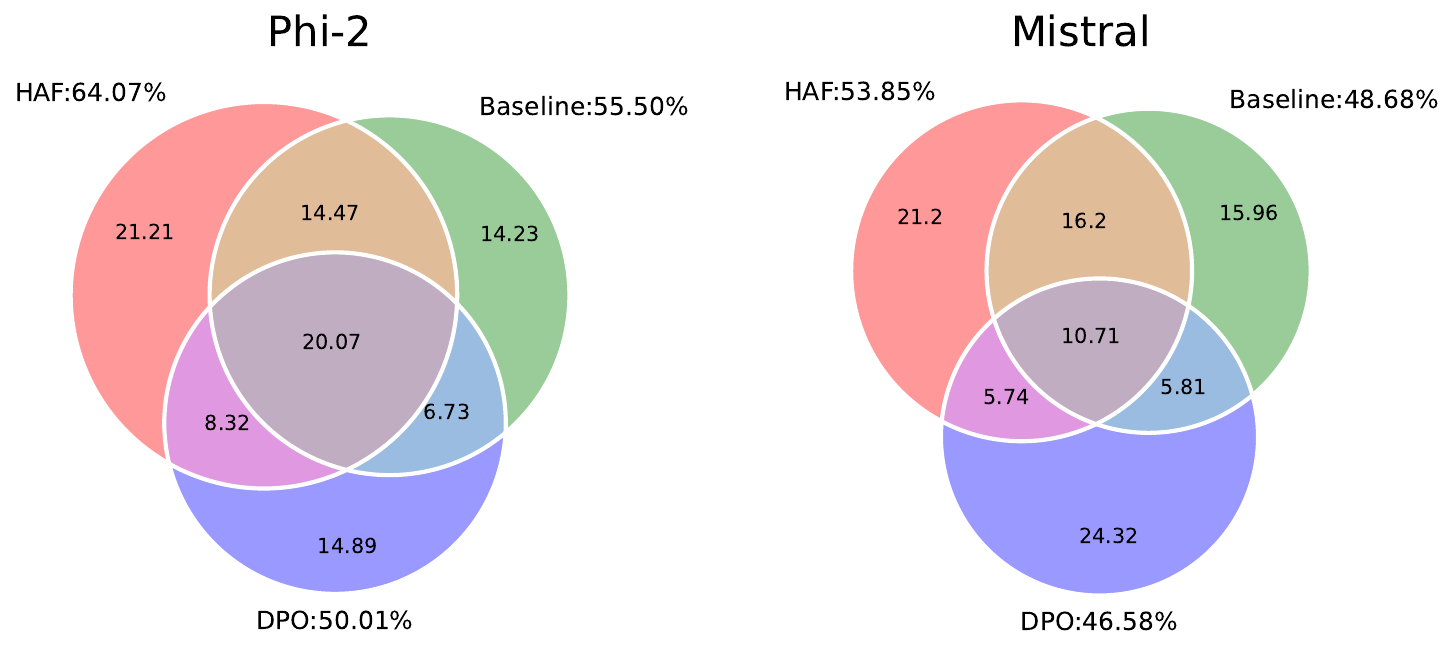}
\caption{Average win rates of responses selected by the HAF reward model, baseline model and the DPO reward model. Circles may overlap as different models select the same response.}
\label{fig:bon}
\end{figure}

We demonstrate the reliability of our trained reward model through Best-of-N selection, where the reward model should pick the best response~(the one with the highest reward) from several responses sampled from the same generative model. The backbone for the reward model and the generation model is the same, with 8 and 4 responses are provided to the Mistral-Instruct reward model and the Phi-2 reward model, respectively. Because Phi-2 tends to generate more similar responses,
reducing the need for 8 candidates. The prompts used for comparisons and ranking are listed in Appendix~\ref{app:gpt-4}, referencing AlpacaEval~\citep{alpaca_eval}.
We report two evaluation metrics. \textbf{Win rate}: We use GPT-4-turbo to rank the responses from \haf, DPO, and baseline reward model and report the win rate~\citep{personalized}. \textbf{Consistency}: we use GPT-4-turbo to rank the sampled responses and calculate the recall of the top-1 and top-2 responses. 



\begin{table}[t]
\centering
\resizebox{0.48\textwidth}{!}{
\resizebox{0.95\linewidth}{!}{
\begin{tabular}{@{}llcccc@{}}\toprule
                       &          & \multicolumn{2}{c}{Phi-2}       & \multicolumn{2}{c}{Mistral}     \\
                       &          & All            & Chat              & All            & Chat              \\\midrule
\multirow{4}{*}{Top-1} & \textit{Random} & \textit{25.00} & \textit{25.00} & \textit{12.50} & \textit{12.50} \\
& Baseline & 27.43          & 28.97          & 16.03          & 18.27          \\
                       & DPO      & 22.94          & 26.39          & 12.81          & 13.85          \\
                       & \haf      & \textbf{33.77} & \textbf{37.19} & \textbf{18.19} & \textbf{21.12} \\\midrule
\multirow{4}{*}{Top-2} & \textit{Random} & \textit{50.00} & \textit{50.00} & \textit{25.00} & \textit{25.00} \\
& Baseline & 49.71          & 53.39          & 30.64          & 35.13          \\
                       & DPO      & 46.22          & 51.59          & 29.05          & 31.56          \\
                       & \haf      & \textbf{58.28} & \textbf{64.23} & \textbf{34.89} & \textbf{39.96}\\\bottomrule
\end{tabular}
}
}
\caption{Top-k recalls of different reward models. \textit{Random} shows the recall when choosing responses randomly. The results are averaged over the recall values from all datasets. ``Chat'' indicates that the result in that column is averaged over the AHP, CA, and Harmless instead of all five datasets. }
\label{tab:consistency}
\end{table}

As shown in Figure~\ref{fig:bon} and Table~\ref{tab:consistency}, \haf demonstrates significant advantages over the baseline and DPO reward models in selecting responses in terms of both evaluation metrics, especially taking Phi-2 as the backbone. Notably, the recall scores of both DPO and baseline are close to those of random selection, indicating poor sensitivity and an inability to discern between responses with minimal quality differences. In contrast, the reward model trained by \haf exhibits good discriminative ability. 

Considering that the model primarily learn to distinguish between harmful and non-harmful responses from the BS and Harmless datasets, and the responses generated by Phi-2 and Mistral are mostly benign, we also report average results on the remaining three datasets. When the safety-related datasets are excluded, all models show an improvement in average performance. The detailed results as well as the ArmoRM-judged results can be found in the appendix in Table~\ref{tab:total_bon}, Figure~\ref{fig:bon_armo}.

Figure~\ref{fig:bon} presents the win rates of each method. We can observe that HAF consistently has the highest probability of selecting the best response (among the three methods), while DPO performs the worst. The frequency with which the baseline reward model and the \haf reward model select the same optimal response is considerably higher than their agreement with DPO. This difference is partly due to their modeling approaches: both \haf and the baseline reward model directly produce numerical rewards, whereas DPO derives rewards from token probabilities. 

\subsubsection{RLHF}\label{sec:rlhf}
We also test \haf in the standard RLHF process: we train two reward models respectively with \haf and the baseline method and then use them to train policy models through RLHF. After training, GPT-4 acts as the evaluator to compare the generations from the two policy models. 
We conduct two sets of experiments: one for training a Safety reward model using the BS and Harmless datasets; and the other for training a Chat reward model using the AHP, CA, and Helpful datasets. We compare the response quality of the policy models optimized after the same number of PPO steps by the baseline reward model and the \haf reward model.

\begin{figure}[t]
\centering
\includegraphics[width=0.37\textwidth]{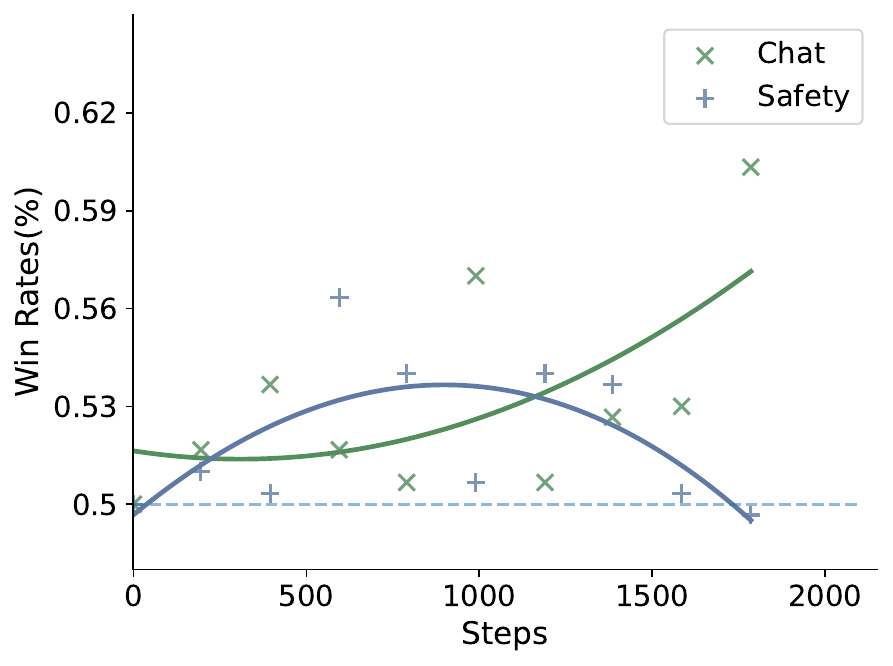}
\caption{Win rates for the policy model trained with the \haf reward model using RLHF compared to the baseline reward model, with each comparison made at the same training steps.}
\label{fig:rlhf}
\end{figure}

As shown in Figure~\ref{fig:rlhf}, the improvement of \haf is particularly evident on the Chat dataset, with its win rate increasing throughout the training, highlighting the superiority of the \haf reward model. In contrast, during safety training, the \haf reward model \textrm{only} shows a significant advantage over the baseline model primarily in the middle stages of training. This is likely because both models have largely achieved harmless responses on the test set, resulting in minimal differentiation between the two reward models.

\section{Related Work}
\label{app:Related Work}

Reward model was proposed to modeling human language preferences (model that outputs preference values based on questions and answers) \citep{DeepRLhumanf}, then the explosive growth of research on reward models~\citep{FragileRM} and large language models~\citep{CoT,GenerativeAI,LLMasaJudge,yue2024synergistic} emerged after the popularity of ChatGPT. 

From training to practical applications, an increasing number of studies have also featured the presence of quantifiable preferences(usually known as ``reward''). For example, RLHF~\citep{DeepRLhumanf,summarizefromHF} uses the PPO algorithm~\citep{PPO} to maximize the reward of the policy model; RAFT~\citep{RAFT} and RRHF~\citep{RRHF} remove substandard data by scoring the candidate responses with reward model; LLM-as-a-judge~\citep{LLMasaJudge} employs GPT-4 to score the text. 

Therefore, how to construct a model offering explicit preference feedback has naturally become a focal point of much research. To train a precise and robust reward model, many studies start from training with human preference data, and many works in the data field are largely centered around this. \citet{llama2} and \citet{SLiC} provided different methods for using ranking data; \citet{secretRLHFRM} explored ways of measuring the strength of the data; while concerning datasets themselves, \citet{IPO}, \citet{Knox2022ModelsOH} and \citet{sensitivity} analyzed the impact of data preference strength on training from theoretical or practical perspectives. In addition, similar to the RAG technique~\citep{RAG} in large language models, many methods~\citep{ToolAugmentedRM,factuallyaugmented} using external tools or references have also emerged, injecting new vitality into the development of reward models. 

Although many data-oriented methods have greatly enhanced the performance of reward models, the field of reward model optimization has been rarely explored. Currently, the training of reward models basically follows the process proposed by OpenAI~\citep{DeepRLhumanf}. 
Considering the widespread practical applications of reward models, the attention given to their training paradigms does not match their importance.

\section{Conclusion}
In this paper, we extend and improve the training framework of the current reward model. We split the training mechanism of the reward model into two stages: aligning model preference and optimizing the reward layer. Through introducing an additional constraint of policy loss, our hybrid alignment framework supervises the internal preference model at the token level while simultaneously optimizing the mapping layer at the sequence level, significantly improving the training effectiveness. We theoretically verify the validity of our method and demonstrate its reliability through systematic experiments. 

Our method allows for a consistent customization of the reward model. In the future, we will thoroughly explore the potential of the reward model and its variants across various tasks, and investigate whether the logistic distribution is the optimal prior for reward modeling.

\section*{Impact Statements}
\label{Impact Statements}
This paper presents work whose goal may benefit the training of large language models in the field of deep learning. Among the many possible consequences, we do not believe that there is a significant possibility of adverse effects on society.
\section*{Limitations}
\label{sec:limitation}
In this paper, we discuss the potential of enhancing the alignment process of reward models by incorporating policy constraints, where the policy loss functions similarly to a regularization loss, acting as an auxiliary function to guide model training. However, since DPO can be directly used to train an implicit reward model, replacing the reward model with a DPO model for downstream tasks can also be a feasible approach, while we do not explore methods for combining the outputs of the policy layer and the reward layer, which remains a direction for our future research.
%


\bibliography{custom}

\appendix


\section{Experiments Setup}
\label{Experiments Setup}
Our default setup is shown in Table~\ref{tab:default setup}. 
\label{app:setup}
\begin{table*}[ht]
\centering
\begin{tabular}{|cc|cc|cc|}
\hline
\textbf{setup} & \textbf{value} & \textbf{setup}         & \textbf{value} & \textbf{setup}               & \textbf{value} \\ \hline
\textbf{lora rank}             & 64           & \textbf{optimizer}     & AdamW          & \textbf{precision}           & bf16           \\
\textbf{lora alpha}                & 16             & \textbf{adam\_beta1}   & 0.9            & \textbf{max gradient norm}   & 1.0            \\
\textbf{training steps}            & 3200           & \textbf{adam\_beta2}   & 0.999          & \textbf{max sequence length} & 512            \\
\textbf{evaluation steps}          & 0.025          & \textbf{weight\_decay} & 0.0            & \textbf{global random seed}  & 0              \\
\textbf{batch size}                & 16             & \textbf{adam\_epsilon} & 1e-5           & \textbf{framework}                 & PyTorch        \\\hline
\end{tabular}
\caption{Default setup}
\label{tab:default setup}
\end{table*}



To train the reward model, we use DPO Loss as the policy loss in HAF and set policy ratio $\alpha=0.2$. The learning rate is $1.0\times10^{-5}$ for Phi-2 and Mistral-lora-baseline, $3.0\times10^{-5}$ for Mistral-lora-HAF. A single RTX A6000 with 48GB memory is used for training the reward model. The model used for testing is the checkpoint that achieves the highest reward on the validation set.

For PPO training in Section~\ref{sec:rlhf}, we utilize two RTX A6000 GPUs for parallel training with a total batch size of 4. The maximum number of new tokens generated is set to 128, and the learning rate is 1e-6. The training is conducted over a maximum of 20,000 episodes. We employ score scaling and score normalization and clip the scores between -3 and 3. All other settings follow the implementation in the TRL library. The model used for testing is the checkpoint that achieves the highest reward on the validation set. The generation config includes $top\_p=0.8$, $temperature=0.5$, $length\_penalty=1.3$, $repetition\_penalty=1.2$, $do\_sample=True$

\section{Discussions for Policy Loss Ratio}
\label{sec:policy loss ratio}
\begin{figure*}
    \centering
    \includegraphics[width=1\textwidth]{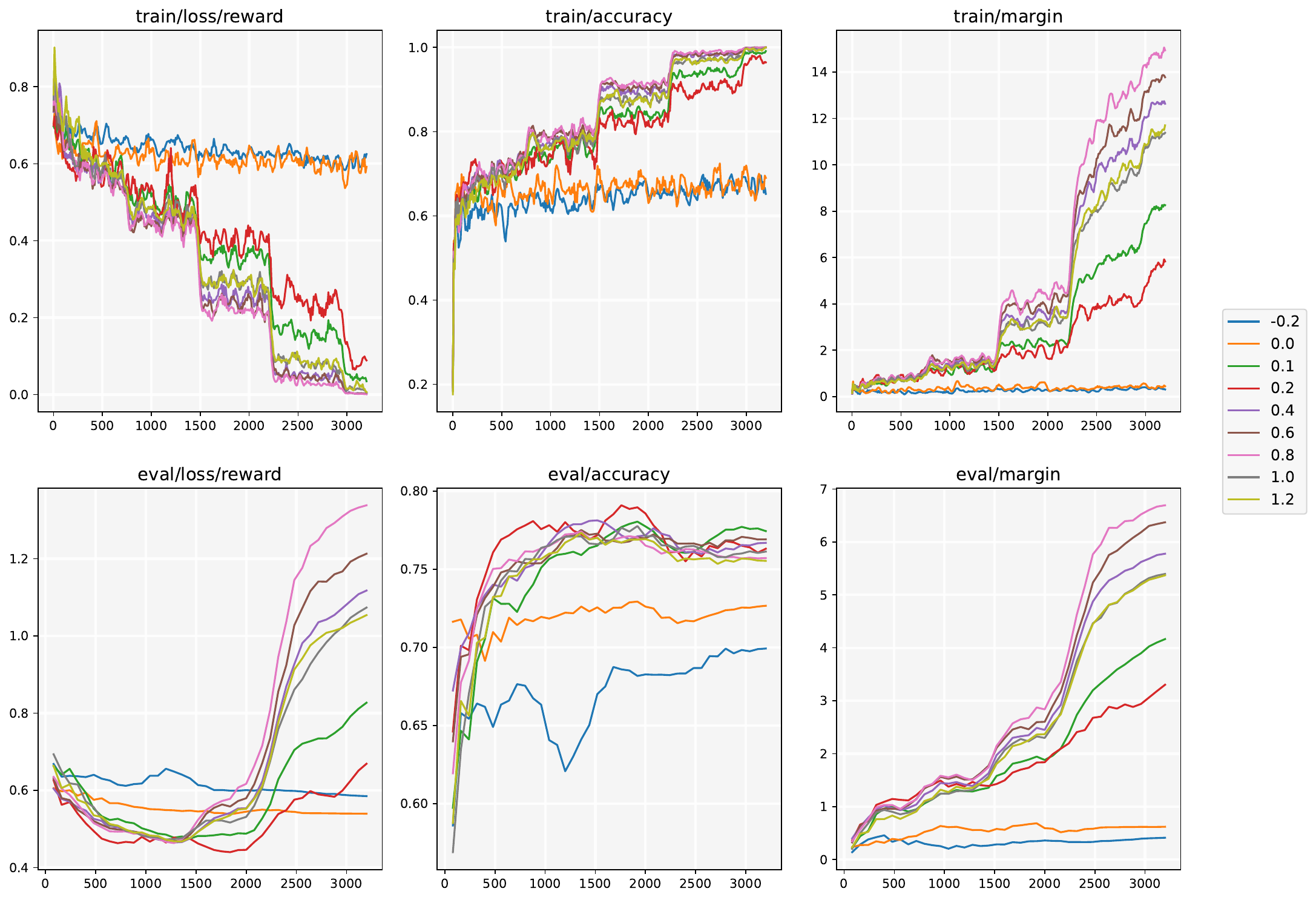}
    \caption{Results for different policy ratios. ``margin'' is the average difference between a better and worse response's rewards. A policy ratio of 0 equals to Baseline method.}
    \label{fig:abl-dpo-ratio}
\end{figure*}
Figure~\ref{fig:abl-dpo-ratio} reveals that incorporating even a mere 0.1x of policy loss can significantly impact the results. Using reward loss alone leads to slow training; to achieve the same loss value, the model with policy loss requires only a fraction of the time. However, this rapid training characteristic also accelerates overfitting, necessitating the use of early stopping strategies to halt training in time. When the policy loss ratio is negative, model performance deteriorates, and the variations in various metrics resemble those of the baseline. This indicates a correlation between the policy model and the reward model.

\section{Loss Functions}
\label{Loss Functions}
\subsection{Deriving the Reward Loss Functions}
\label{app:reward loss}

In the Bradley-Terry model's assumption, Oracle reward model outputs rewards in connection with the win rates:
\begin{equation}
    \underset{p\sim J}{\mathbb{E}}\mathbb{I}(y>y';x,p)=-\log\sigma[\bm{r}^*(x,y)-\bm{r}^*(x,y')]
\label{eq:essential assumption}
\end{equation}
where $p$ is a judge~(annotator) sampled from the judge distribution $J$.

As we only focus on the reward differences between responses to the same prompt, there exists another metric denoted as $\mathrm{D}_1'$ for calculating the reward loss:
\begin{equation*}
\begin{aligned}
    \mathcal{L}_s=\underset{x,y,y'}{\mathbb{E}}\mathrm{D}_1'[&r(x,y)-r(x,y'),\\&\quad r^*(x,y)-r^*(x,y')]
\end{aligned}
\end{equation*}

As $-\log\sigma(\cdot)$ is monotonically increasing, so there exists a metric $\mathrm{D}_1''$, such that
\begin{equation*}
\begin{aligned}
    &\mathrm{D}_1'[r(x,y)-r(x,y'),r^*(x,y)-r^*(x,y')]\\
    =&\mathrm{D}_1''[-\log\sigma(\bm{r}(x,y)-\bm{r}(x,y')),\\&\qquad\qquad\qquad\ -\log\sigma(\bm{r}^*(x,y)-\bm{r}^*(x,y'))]\\
    =&\mathrm{D}_1''[-\log\sigma(\bm{r}(x,y)-\bm{r}(x,y')),\\&\qquad\qquad\qquad\qquad\qquad \underset{p\sim J}{\mathbb{E}}\mathbb{I}(y>y';x,p)]
\end{aligned}
\end{equation*}

Let $\mathrm{D}_1''$ be the cross-entropy loss, and let $\mathrm{P}(x,y,y')=-\log\sigma(\bm{r}(x,y)-\bm{r}(x,y'))$,
\begin{equation*}
\begin{aligned}
    &\mathcal{L}_s=\underset{x,y,y'}{\mathbb{E}}[\mathrm{P}(x,y,y')\cdot\underset{p\sim J}{\mathbb{E}}\mathbb{I}(y>y';x,p)\\&\quad+(1-\mathrm{P}(x,y,y'))\cdot(1-\underset{p\sim J}{\mathbb{E}}\mathbb{I}(y>y';x,p))]\\
    &=\underset{p\sim J}{\underset{x,y,y'}{\mathbb{E}}}[\mathrm{P}(x,y,y')\cdot\mathbb{I}(y>y';x,p)\\&\quad+(1-\mathrm{P}(x,y,y'))\cdot(1-\mathbb{I}(y>y';x,p))]
\end{aligned}
\end{equation*}
which is exactly Eq.~\ref{eq:standard_reward} when we sample from $\mathcal{D}$.
\subsection{DPO as the Policy Loss}
\label{app:dpo as policy loss}
The derivation for policy loss is the same as reward loss in their essence. The policy model can be treated as a reward model with sequence probabilities reflecting the rewards~\citep{DPO,DPOQFunc}. $reward(x,y)=\log[\pi(x,y)/\pi_{ref}(x,y)]$. 

From this perspective, the DPO loss and reward loss share the same assumption~(Eq.~\ref{eq:essential assumption}). The reward model and the DPO-trained policy model are essentially doing the same task despite some formal differences~\citep{DPO,DPOQFunc}.

\section{Mathematical Enlightenment}
\label{app:math enlight}

\subsection{Theoretical Explanation for the Claims}
\label{app:theory-explain}
\paragraph{Inequality for claim 1.}
Unless $K$ can exactly fit $K^*$, there exists $\epsilon>0$, such that
\begin{equation*}
\begin{aligned}
    &\quad\underset{d\sim \mathcal P}{\mathbb E}[\mathrm D_2(\mathrm{K_H\circ\phi_H}(d),\mathrm{K^*\circ\phi^*}(d))]\\&\leqslant \underset{\mathrm K}{\mathrm{min}}\underset{d\sim \mathcal P}{\mathbb E}[\mathrm D_2(\mathrm{K\circ\phi_s}(d),\mathrm{K^*\circ\phi^*}(d))]-\frac{\epsilon}{\alpha}
\end{aligned}
\label{f}
\end{equation*}
\noindent holds for all $\alpha\in(0.1,2)$, where $K_H, \phi_H = \underset{\mathrm K, \phi}{\mathrm{argmin}}\mathcal{L}_H$ in Equation~\ref{eq:calibrated_reward} and $\phi_s = \underset{\phi}{\mathrm{argmin}}\mathcal{L}_{s}$ in Equation~\ref{eq:standard_reward}. 
Here we use $\mathrm{argmin}$ to represent the best models optimized with the corresponding loss functions, so $\phi_H$ and $\phi_s$ are not equal to $\phi^*$ although $\phi^*$ is the minimum mathematically. 

\paragraph{Inequality for claim 2.}
Assume that $\phi^*$ is unique, $K^*$ is locally Lipschitz continuous, 
, and $0.1<\alpha<2$, there exists $k,\delta>0$, such that
\begin{equation*}
\begin{aligned}    &\underset{d\sim\mathcal P}{\mathbb{E}}[|\phi_H(d)-\phi^*(d)|-|\phi_s(d)-\phi^*(d)|]\nonumber<\\&\frac{g_{\max}-g_{\min}}{g_{\min}}\underset{d\sim\mathcal P}{\mathbb{E}}|\phi_s(d)-\phi^*(d)|+2\delta-\frac \epsilon {\alpha\cdot k}
\end{aligned}
\label{g}
\end{equation*}

We obtain informally here an upper bound on the model preference error. By tuning the hyperparameter $\alpha$, the right term can be strictly negative.

\subsection{Inequality Scaling}

\begin{align*}
\mathop{\min}_{\mathrm{F,\phi,K}}\;\mathop{\mathbb{E}}_{d\sim\mathcal P}[&\mathrm D_1(\mathrm{F\circ\phi}(d),\mathrm{F^*\circ\phi^*}(d))\\+\alpha\cdot&\mathrm D_2(\mathrm{K\circ\phi}(d),\mathrm{K^*\circ\phi^*}(d))]\\\leqslant \;\underset{\mathrm{\underset{K}{\underset{\phi=\phi_s}{F=F_s}}}}{\min}\;\underset{d\sim \mathcal P}{\mathbb E}[&\mathrm D_1(\mathrm{F\circ\phi}(d),\mathrm{F^*\circ\phi^*}(d))\\+\alpha\cdot&\mathcal L_2(\mathrm{K\circ\phi}(d),\mathrm{K^*\circ\phi^*}(d))]\\
=\;\underset{\mathrm{K}}{\min}\;\underset{d\sim \mathcal P}{\mathbb E}[&\alpha\cdot\mathrm D_2(\mathrm{K\circ\phi_s}(d),\mathrm{K^*\circ\phi^*}(d))]\\+\underset{d\sim \mathcal P}{\mathbb E}&[\mathrm D_1(\mathrm{F_s\circ\phi_s}(d),\mathrm{F^*\circ\phi^*}(d))]
\end{align*}

\noindent With the definition of $\phi_H,\mathrm K_H,\mathrm F_H$, we have:
\begin{align*}
&\underset{d\sim \mathcal P}{\mathbb E}[\mathrm D_1(\mathrm{F_H\circ\phi_H}(d),\mathrm{F^*\circ\phi^*}(d))\\&\qquad+\alpha\cdot\mathrm D_2(\mathrm{K_H\circ\phi_H}(d),\mathrm{K^*\circ\phi^*}(d))]\\\leqslant &\underset{d\sim \mathcal P}{\mathbb E}[\mathrm D_1(\mathrm{F_s\circ\phi_s}(d),\mathrm{F^*\circ\phi^*}(d))]\\&\qquad+\underset{\mathrm{K}}{\min}\underset{d\sim \mathcal P}{\mathbb E}[\alpha\cdot\mathrm D_2(\mathrm{K\circ\phi_s}(d),\mathrm{K^*\circ\phi^*}(d))]\\\leqslant &\underset{d\sim \mathcal P}{\mathbb E}[\mathrm D_1(\mathrm{F_H\circ\phi_H}(d),\mathrm{F^*\circ\phi^*}(d))]\\&\qquad+\underset{\mathrm{K}}{\min}\underset{d\sim \mathcal P}{\mathbb E}[\alpha\cdot\mathrm D_2(\mathrm{K\circ\phi_s}(d),\mathrm{K^*\circ\phi^*}(d))]
\end{align*}

\noindent In practical settings, ``$\leqslant$''s do not hold at the same time~(simultaneously optimizing two objectives is preferable to optimizing them sequentially). With the premise that the model is fully optimized with the hybrid alignment loss for any $\alpha \in(0.1,2)$, which means both of the objectives have an impact on the final optimization result, namely $\phi_H\neq \phi_s$, there exists a little gap $\epsilon> 0$ such that
\begin{align*}
&\underset{d\sim \mathcal P}{\mathbb E}[\mathrm D_1(\mathrm{F_H\circ\phi_H}(d),\mathrm{F^*\circ\phi^*}(d))\\&\qquad+\alpha\cdot\mathrm D_2(\mathrm{K_H\circ\phi_H}(d),\mathrm{K^*\circ\phi^*}(d))]\\
\leqslant&\underset{d\sim \mathcal P}{\mathbb E}[\mathrm D_1(\mathrm{F_H\circ\phi_H}(d),\mathrm{F^*\circ\phi^*}(d))]\\&+\underset{\mathrm{K}}{\min}\underset{d\sim \mathcal P}{\mathbb E}[\alpha\cdot\mathrm D_2(\mathrm{K\circ\phi_s}(d),\mathrm{K^*\circ\phi^*}(d))]-\epsilon 
\end{align*}
Then, there goes
\begin{equation*}
\begin{aligned}
    &\quad\underset{d\sim \mathcal P}{\mathbb E}[\mathrm D_2(\mathrm{K_H\circ\phi_H}(d),\mathrm{K^*\circ\phi^*}(d))]\\&\leqslant \underset{\mathrm{K}}{\min}\underset{d\sim \mathcal P}{\mathbb E}[\mathrm D_2(\mathrm{K\circ\phi_s}(d),\mathrm{K^*\circ\phi^*}(d))]-\frac{\epsilon}{\alpha}\phantom{11111111}
\end{aligned}
\end{equation*}
Here we get the first inequality.

\subsection{Derive the Final Inequality with the 3 Properties}
\label{app:derivation}
\textbf{Convergence}:

Since the trained model $\mathrm K\circ\phi$ is close to $\mathrm K^*\circ\phi^*$, we can therefore linearize $\mathrm D_2$ with a certain positive number $k$:
\begin{equation}
\begin{aligned}
    &\underset{d\sim\mathcal P}{\mathbb{E}} [\mathrm D_2(\mathrm{K}\circ\phi(d),\mathrm{K}^*\circ\phi^*(d))]\\= &\underset{d\sim\mathcal P}{\mathbb{E}}k|\mathrm{K}\circ\phi(d)-\mathrm{K}^*\circ\phi^*(d)|
\end{aligned}
\label{convergence}
\end{equation}\\
\textbf{Separating little disturbance}:
\begin{equation}
    \underset{d\sim\mathcal P}{\mathbb{E}}|\mathrm N\circ\phi(d)|< \delta
\label{disturbance}
\end{equation} holds for any fully-optimized model $\mathrm K\circ\phi$ with $\mathrm N\coloneqq\mathrm K-\mathrm K^*$. Given that the trained model and its preferences closely approximate those of the true model and preferences, we are able to scale down the error terms by a small margin.

\noindent\textbf{Gradient scaling}: 

Intuitively, the optimal model is unique, so $\underset{d\sim\mathcal P}{\mathbb E}|\mathrm K^*\circ\phi(d)-\mathrm K^*\circ\phi^*(d)|>0$. Here we make a slightly stronger assumption that $\mathrm K^*$ is locally $g_{max}$-Lipschitz continuous and has the lower bound $g_{min}$, which means for any $\phi$ that is close to $\phi^*$, there exists
\begin{equation}
\begin{aligned}
    &g_{min}\underset{d\sim\mathcal P}{\mathbb E}||\phi(d)-\phi^*(d)||\\
    <&\underset{d\sim\mathcal P}{\mathbb E}|\mathrm K^*\circ\phi(d)-\mathrm K^*\circ\phi^*(d)|\\
    <&g_{max}\underset{d\sim\mathcal P}{\mathbb E}||\phi(d)-\phi^*(d)||
    \label{gradient}
\end{aligned}
\end{equation}

Based on these three properties, we can derive the result from Appendix~\ref{f}.
\begin{align}
    &\textbf{Inequality 2}\nonumber\\
    &\stackrel {\text{Eq.}~\ref{convergence}}\Longrightarrow \underset{d\sim\mathcal P}{\mathbb{E}}|\mathrm{K_H}\circ\phi_H(d)-\mathrm{K}^*\circ\phi^*(d)|\nonumber\\&\quad\leqslant\underset{\mathrm{K}}{\min}\underset{d\sim\mathcal P}{\mathbb{E}}|\mathrm{K}\circ\phi_s(d)-\mathrm{K}^*\circ\phi^*(d)|-\frac {\epsilon}{\alpha\cdot k}\nonumber\\&\stackrel {\text{Ineq.}~\ref{disturbance}}\Longrightarrow \underset{d\sim\mathcal P}{\mathbb{E}} |\mathrm{K^*}\circ\phi_H(d)-\mathrm{K}^*\circ\phi^*(d)|-\delta\nonumber\\&\quad< \underset{d\sim\mathcal P}{\mathbb{E}}|\mathrm{K^*}\circ\phi_s(d)-\mathrm{K}^*\circ\phi^*(d)|+\delta-\frac{\epsilon}{\alpha\cdot k}\nonumber\\&\stackrel {\text{Ineq.}~\ref{gradient}}\Longrightarrow \nonumber\\&\;g_{min}\underset{d\sim\mathcal P}{\mathbb{E}}[||\phi_H(d)-\phi^*(d)||-||\phi_s(d)-\phi^*(d)||]\nonumber\\&\,<(g_{max}-g_{min})\underset{d\sim\mathcal P}{\mathbb{E}}||\phi_s(d)-\phi^*(d)||\nonumber\\&\qquad\qquad\qquad\qquad\qquad\qquad\qquad+2\delta-\frac \epsilon {\alpha\cdot k}\nonumber
\label{result}
\end{align}

\section{Experiment Results}
\subsection{Consistency}
\label{app:consistency}

\begin{figure}[t]
\centering
\includegraphics[width=0.45\textwidth]{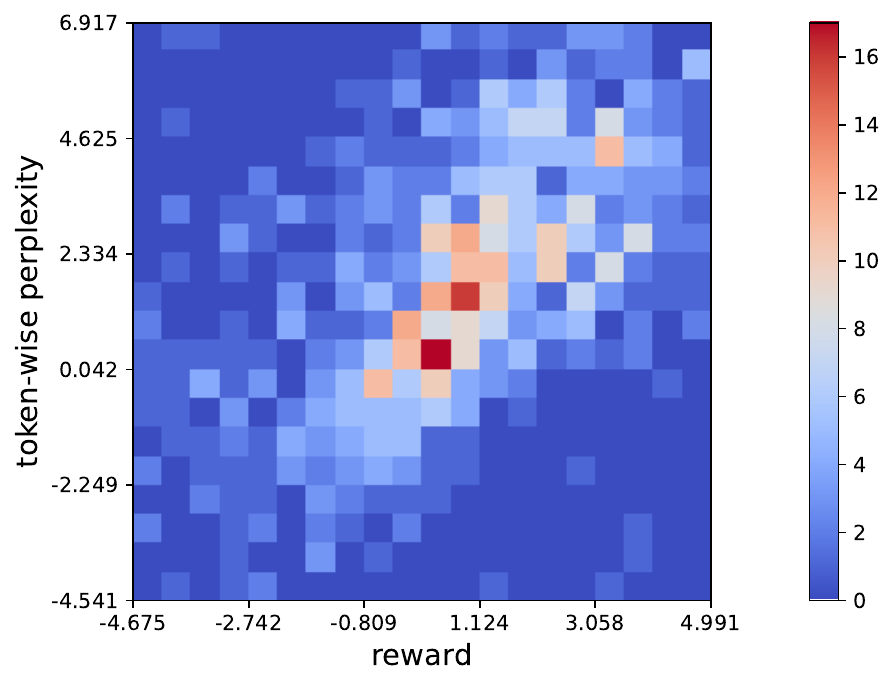}
\caption{The distribution of the reward model's and DPO model's outputs on test data when trained with identical data.}
\label{fig:reward_ppl_heat}
\end{figure}

The x-axis in Figure~\ref{fig:consistency} represents the reward difference between the responses generated by the DPO model and those generated by the HAF model's policy head. This difference is scored by the reward model trained on the same data distribution, which we refer to as the Oracle reward model. We retain the checkpoints from the training processes of both DPO and \haf model and identify potential model pairs with similar performance using five methods (corresponding to the five colors in the figure). This similarity in performance ensures that the higher reward is not a result of better response quality. The five methods include ``reward'' (similar scores from the Oracle reward model), ``acc'' (similar binary classification accuracy), ``loss'' (similar loss values), ``margin'' (similar average margins of model predictions), and ``step'' (same training steps).

It can be observed that the differences in \haf scores are almost always higher than those from the Oracle reward model. This suggests that the preferences of the reward model are influenced by the preferences of the shared parameter policy model, providing some evidence for the existence of an Internal Preference Model.

Also shown in Figure~\ref{fig:reward_ppl_heat}, we independently trained a DPO model and a reward model using the same data and observed a strong positive correlation (even linearity) in their predictions on the test data. This indicates a significant similarity in the preference modeling processes of the DPO model and the reward model. A response preferred by the reward model will also be preferred by the DPO model, which we introduce the concept of the ``Internal Preference Model'' to explain.

\subsection{Overall Performance}
\label{app:overallperformance}

\begin{table*}[t]
\centering
\begin{tabular}{@{}clcccccc@{}}
\toprule
Model&\multicolumn{1}{c}{Metric}& \makecell{Helpful} & \makecell{Harmless} & \makecell{CA} & \makecell{BS} & \makecell{AHP} &\\ 
\midrule
\multirow{2}{*}{Phi-2}& pp$_{win}$ & 0.74 & 1.00 & \textbf{0.60} & 0.74 & 2.52 \\
& pp$_{lose}$ & 0.92 & 0.97 & 1.09 & \textbf{0.60} & 2.55 \\
\multirow{2}{*}{Mistral-base}& pp$_{win}$ & 0.42 & 0.65 & 0.51 & \textbf{0.38} & 0.75 & \\
& pp$_{lose}$ & 0.62 & 0.63 & 0.87 & \textbf{0.28} & 0.98 & \\
\multirow{2}{*}{Mistral-Instruct}& pp$_{win}$ & 3.50 & 5.13 & 2.33 & \textbf{1.58} & 1.98 \\
& pp$_{lose}$ & 6.08 & 5.81 & 3.52 & \textbf{1.31} & 2.67 \\
 \bottomrule
\end{tabular}
\caption{Variances of the corresponding metrics. ``pp'' means token-wise perplexity. The subscript ``win'' refers to the better response while ``lose'' refers to the worse response.}
\label{tab:mean perplex}
\end{table*}

\begin{table}[t]
\centering
\vspace{2mm}
\setlength{\abovecaptionskip}{1mm}
\setlength{\belowcaptionskip}{1mm}{
\begin{tabular}{@{}lccc@{}}\toprule
&pp$_{win}$&pp$_{lose}$&pp$_{win}$-pp$_{lose}$\\ \midrule
$corr$& $-0.8166$ & $-0.9492$ & $-0.9064$\\
$p$& $0.0916$ & $0.0136$ & $0.0339$\\\bottomrule
\end{tabular}
}
\caption{The Pearson correlation coefficient between the variance of the token-wise perplexity of Mistral-Instruct and the difference in accuracy between the reward model trained with DPO and the accuracy of the baseline training. ``corr'' indicates the Pearson correlation coefficient, while ``p'' indicates significance.}
\label{tab:person corr}
\end{table}

Table~\ref{tab:mean perplex} shows the token-wise perplexity calculated by each model for each dataset. $$pp=-\frac{\log \mathrm{Prob}(sequence)}{\mathrm{Length}(sequence)}$$

Another interesting finding is that the variance of the token-wise perplexity (pp) values for Mistral-Instruct shows a very strong negative correlation with the performance of the DPO reward model. Table~\ref{tab:person corr} calculates the Pearson correlation coefficient between the variance of the pp values and the performance difference between the DPO reward model and the baseline reward model, indicating that this negative correlation is highly significant. This finding may provide valuable insights for aligning well-trained (but not yet well-aligned) models.

\subsection{Best of N}
\begin{figure*}[htbp]
    \centering
    \begin{subfigure}[b]{0.45\textwidth}
        \centering
        \includegraphics[width=\textwidth]{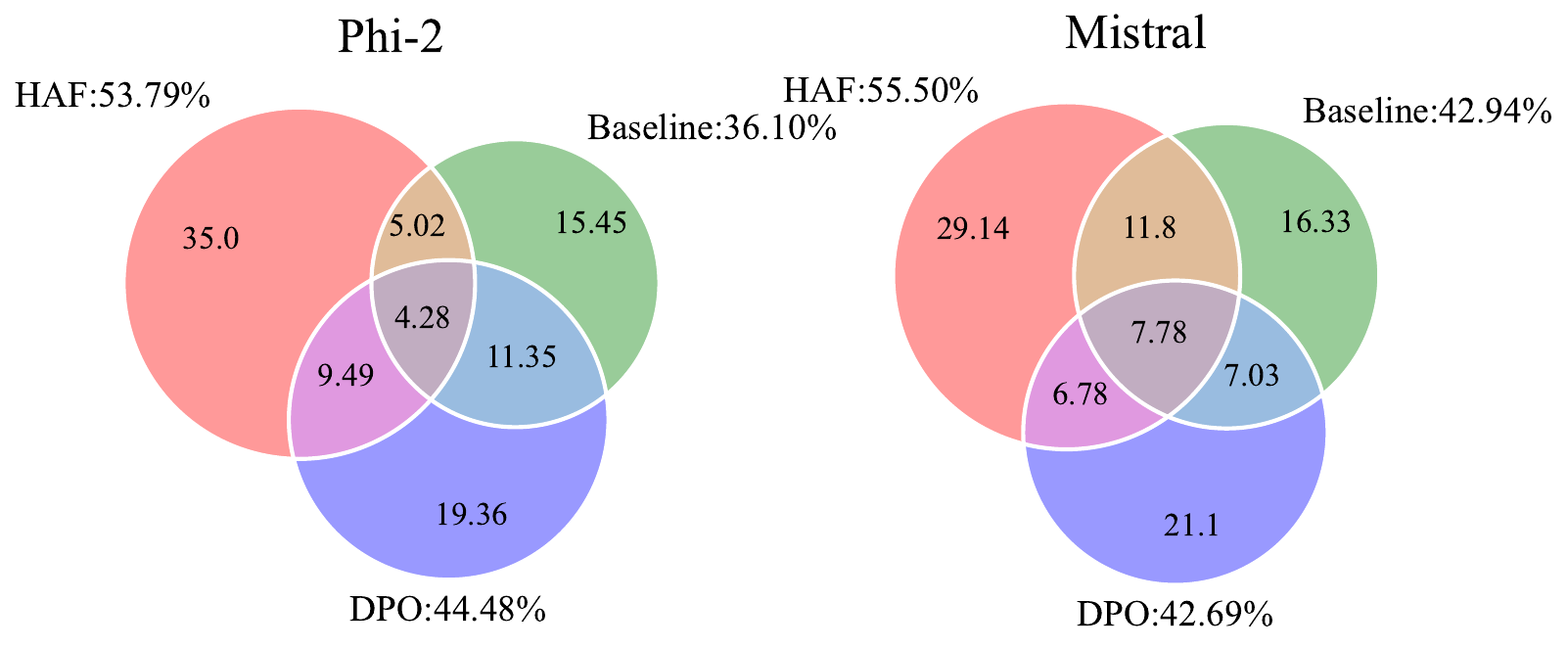}
        \caption{AHP}
    \end{subfigure}
    \begin{subfigure}[b]{0.45\textwidth}
        \centering
        \includegraphics[width=\textwidth]{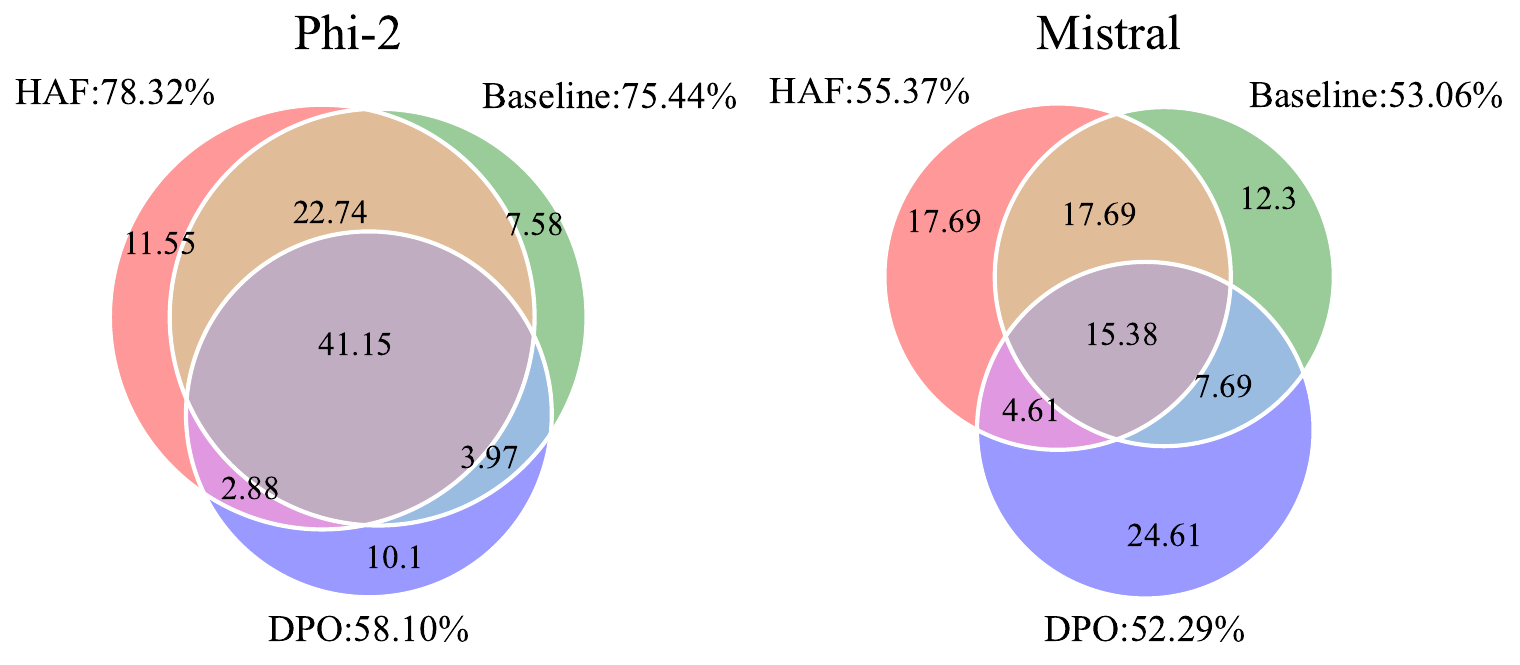}
        \caption{BS}
    \end{subfigure}
    
    \vspace{1em}
    
    \begin{subfigure}[b]{0.45\textwidth}
        \centering
        \includegraphics[width=\textwidth]{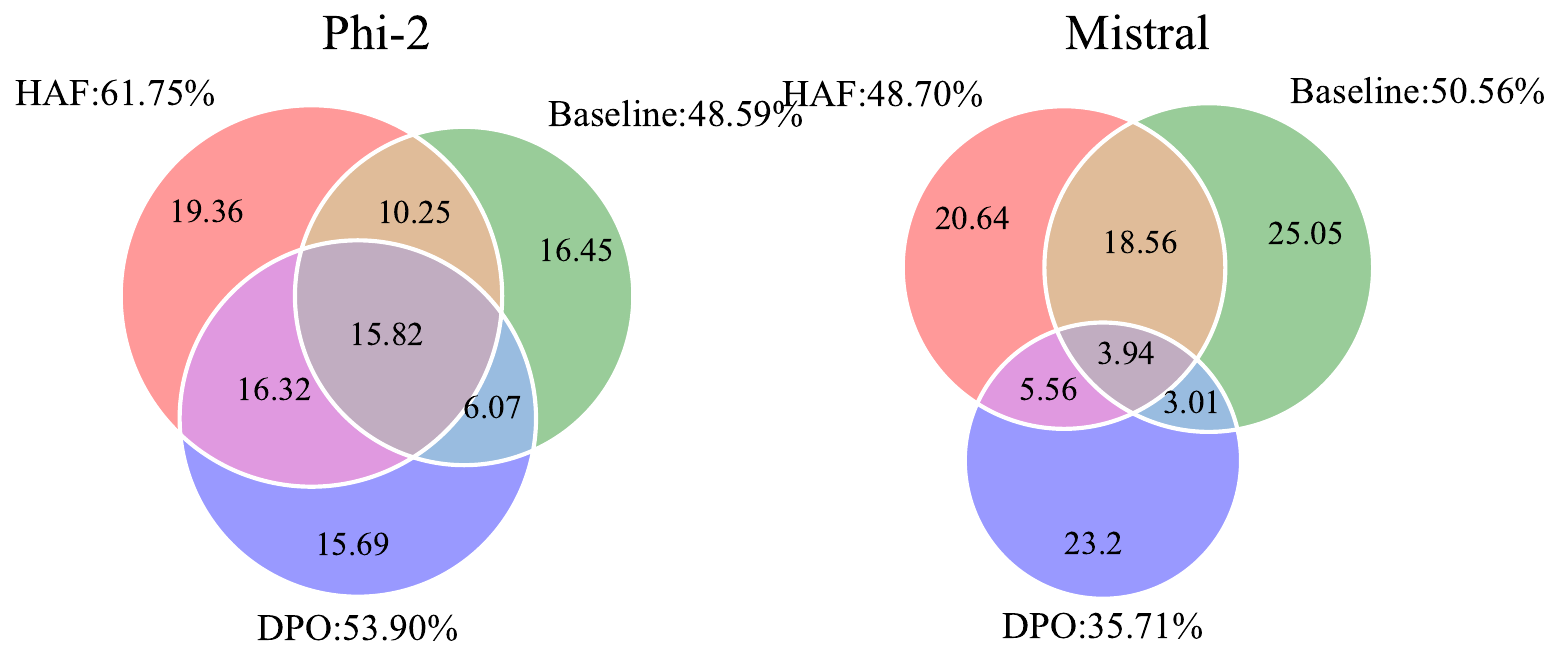}
        \caption{CA}
    \end{subfigure}
    \begin{subfigure}[b]{0.45\textwidth}
        \centering
        \includegraphics[width=\textwidth]{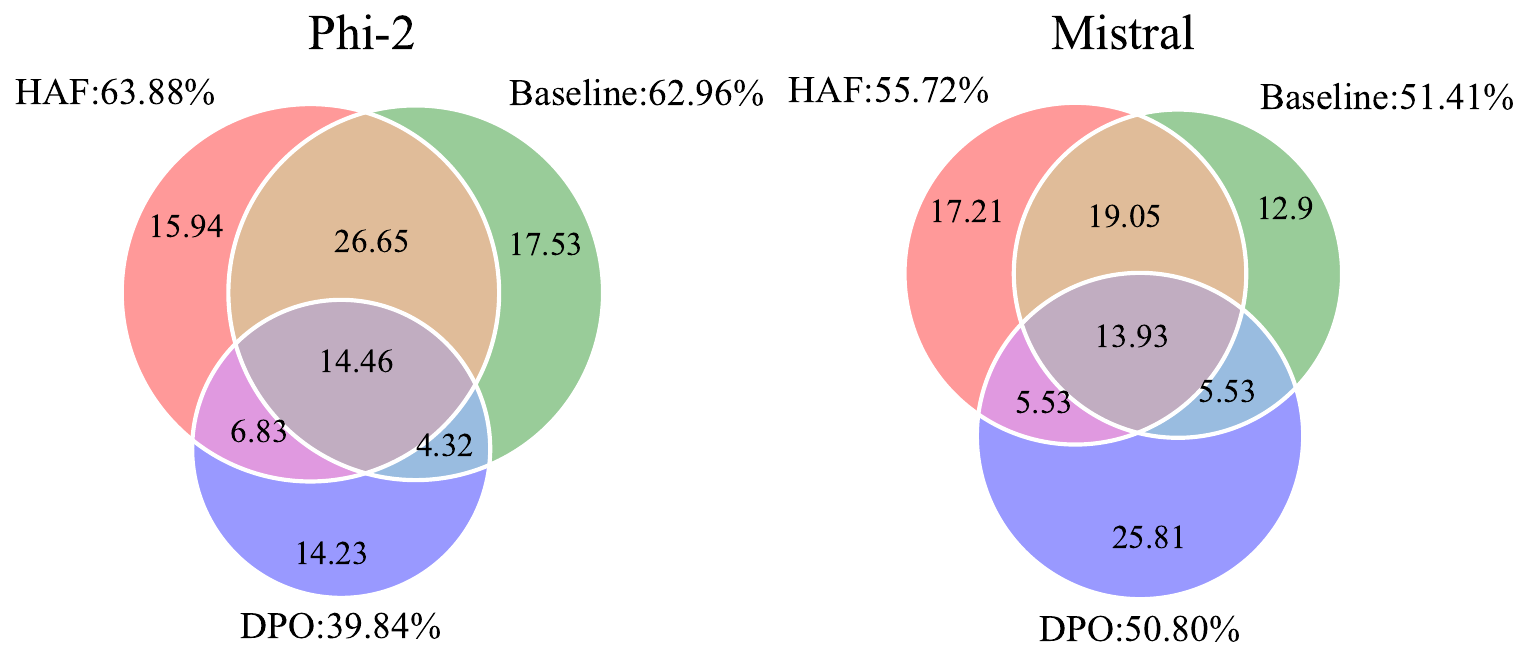}
        \caption{Helpful}
    \end{subfigure}
    
    \vspace{1em}
    
    \begin{subfigure}[b]{0.45\textwidth}
        \centering
        \includegraphics[width=\textwidth]{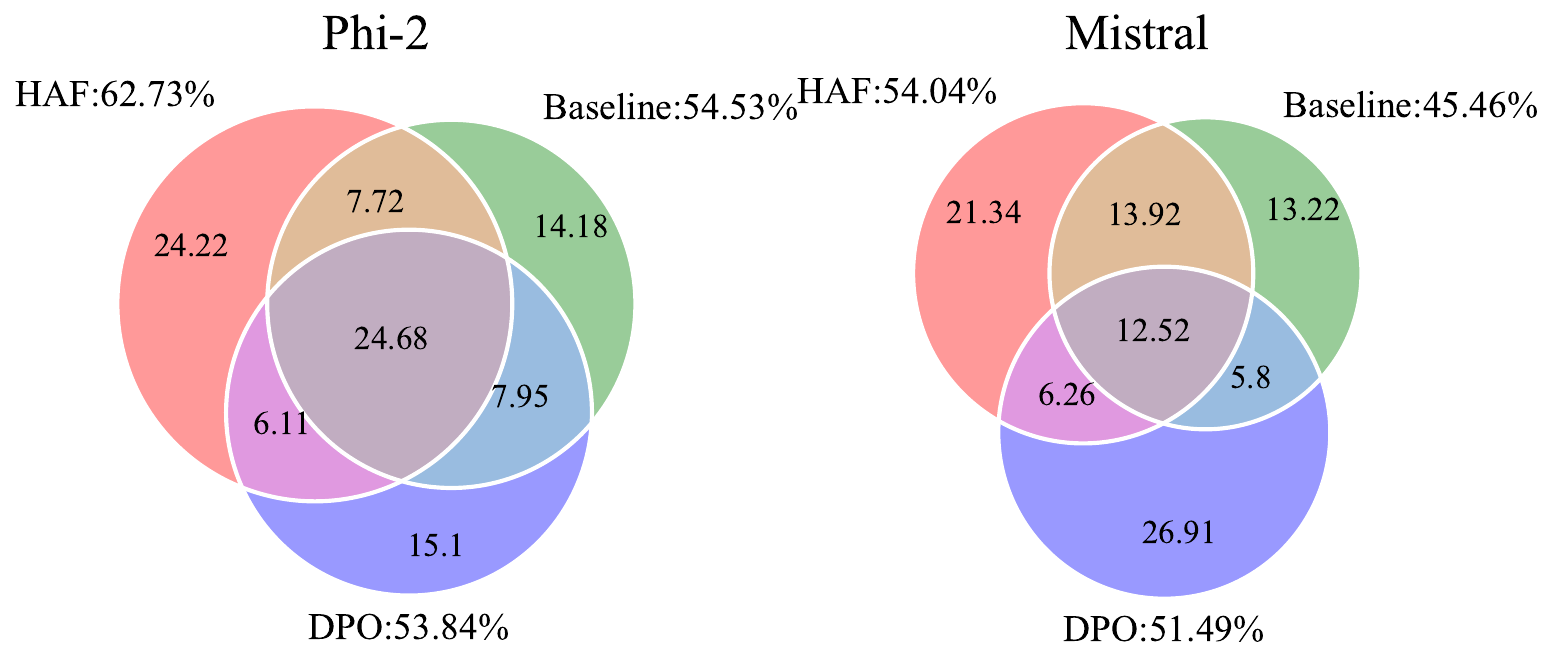}
        \caption{Harmless}
    \end{subfigure}

    \caption{Win rates on each dataset judged by \texttt{gpt-4-turbo-2024-04-09}}
    \label{fig:detailed-venn}
\end{figure*}

\begin{figure*}[htbp]
    \centering
    \begin{subfigure}[b]{0.45\textwidth}
        \centering
        \includegraphics[width=\textwidth]{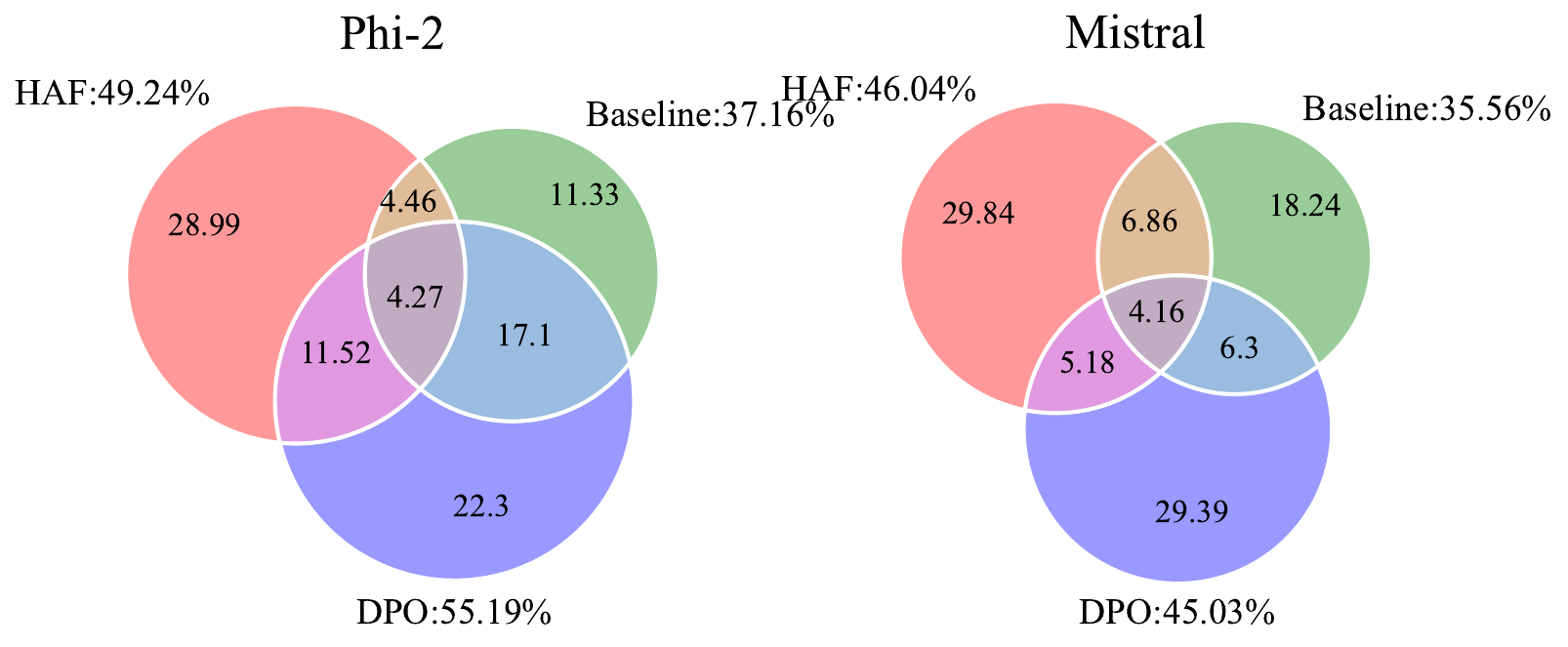}
        \caption{AHP}
    \end{subfigure}
    \begin{subfigure}[b]{0.45\textwidth}
        \centering
        \includegraphics[width=\textwidth]{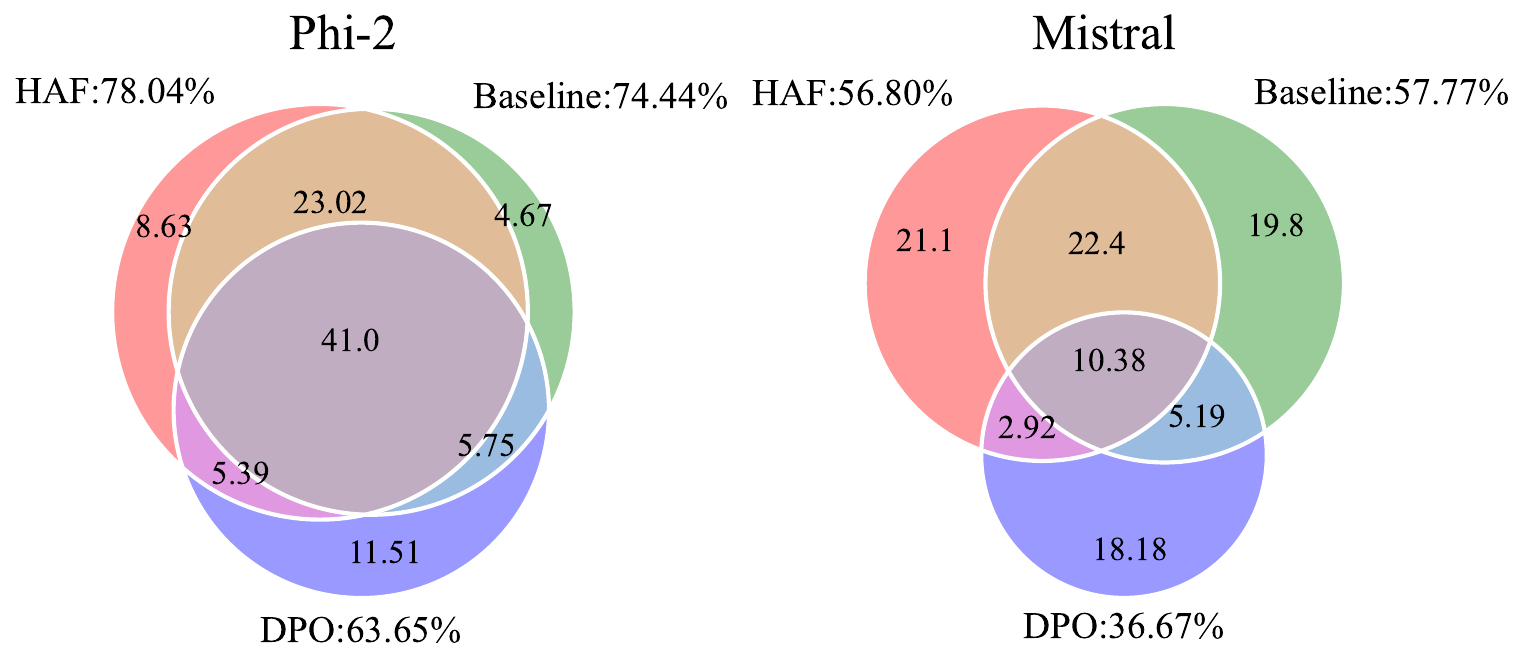}
        \caption{BS}
    \end{subfigure}
    
    \vspace{1em}
    
    \begin{subfigure}[b]{0.45\textwidth}
        \centering
        \includegraphics[width=\textwidth]{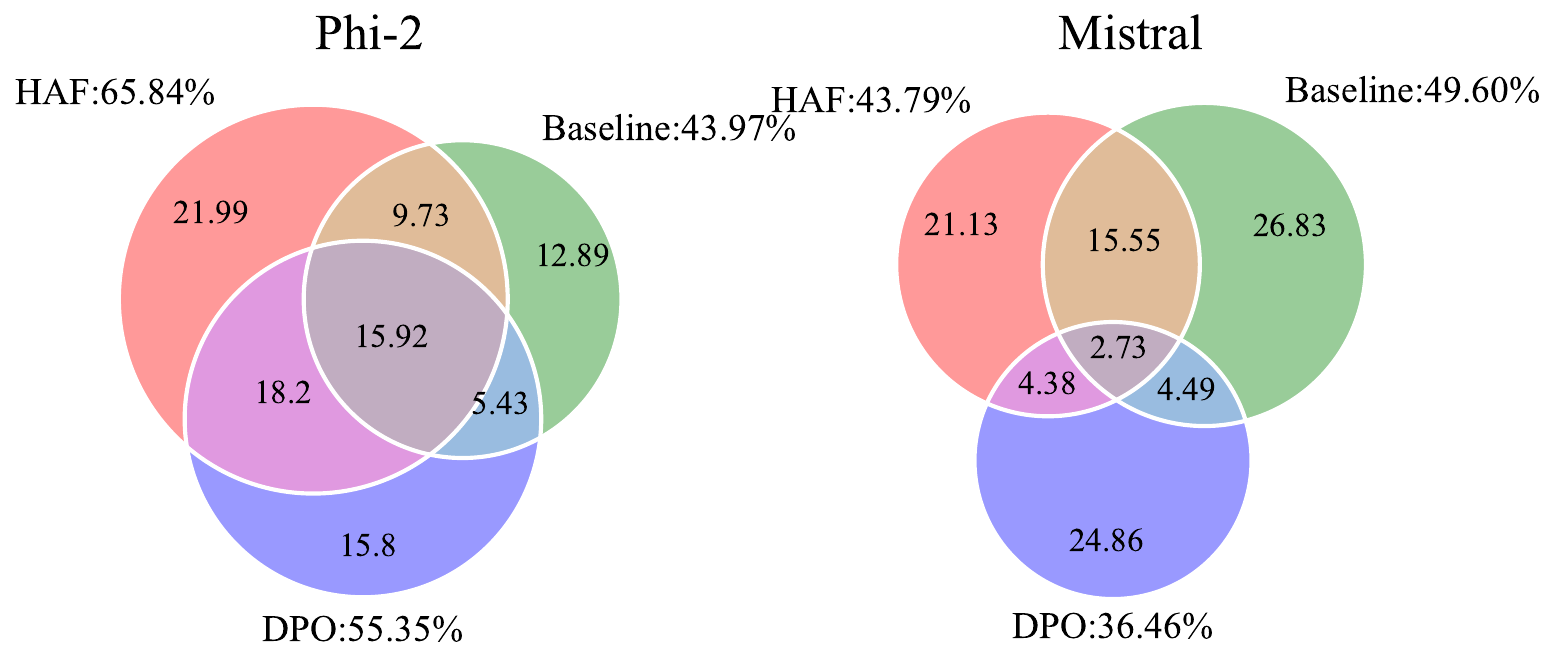}
        \caption{CA}
    \end{subfigure}
    \begin{subfigure}[b]{0.45\textwidth}
        \centering
        \includegraphics[width=\textwidth]{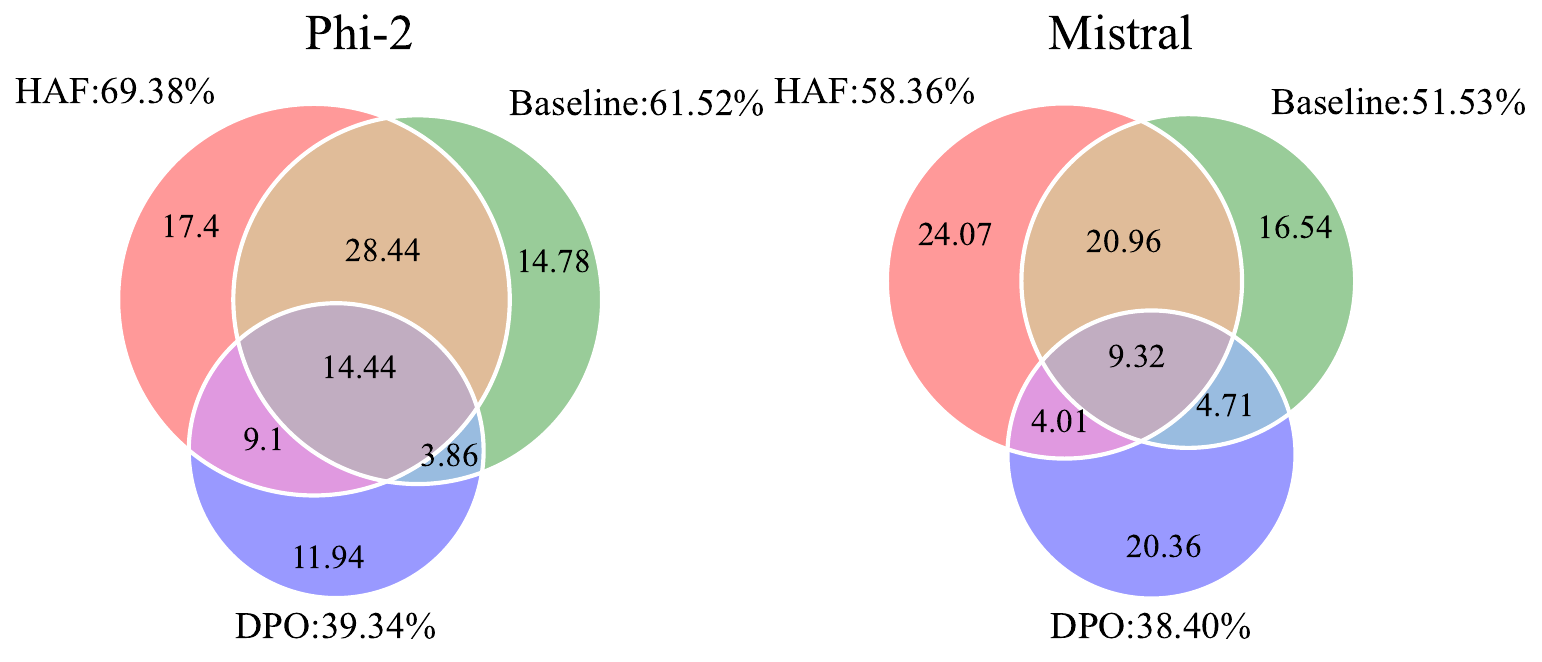}
        \caption{Helpful}
    \end{subfigure}
    
    \vspace{1em}
    
    \begin{subfigure}[b]{0.45\textwidth}
        \centering
        \includegraphics[width=\textwidth]{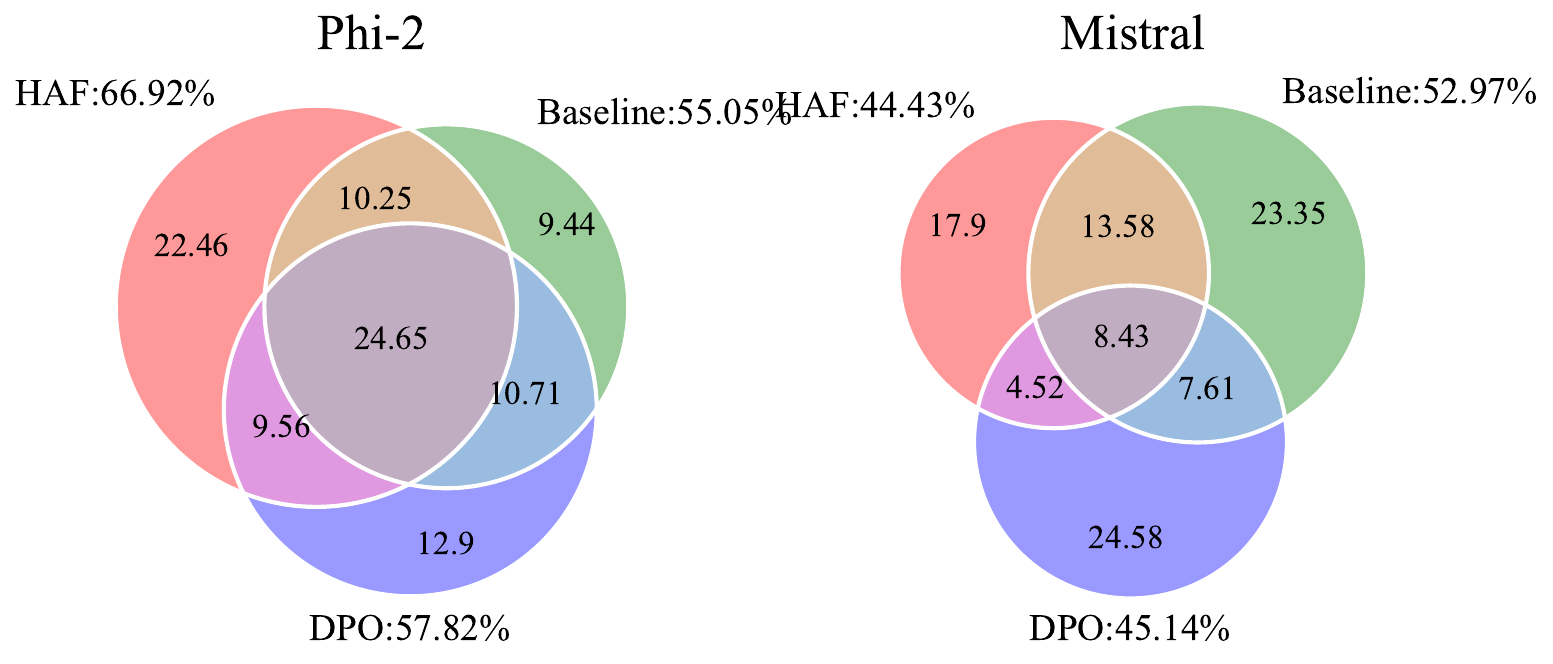}
        \caption{Harmless}
    \end{subfigure}
    \begin{subfigure}[b]{0.45\textwidth}
        \centering
        \includegraphics[width=\textwidth]{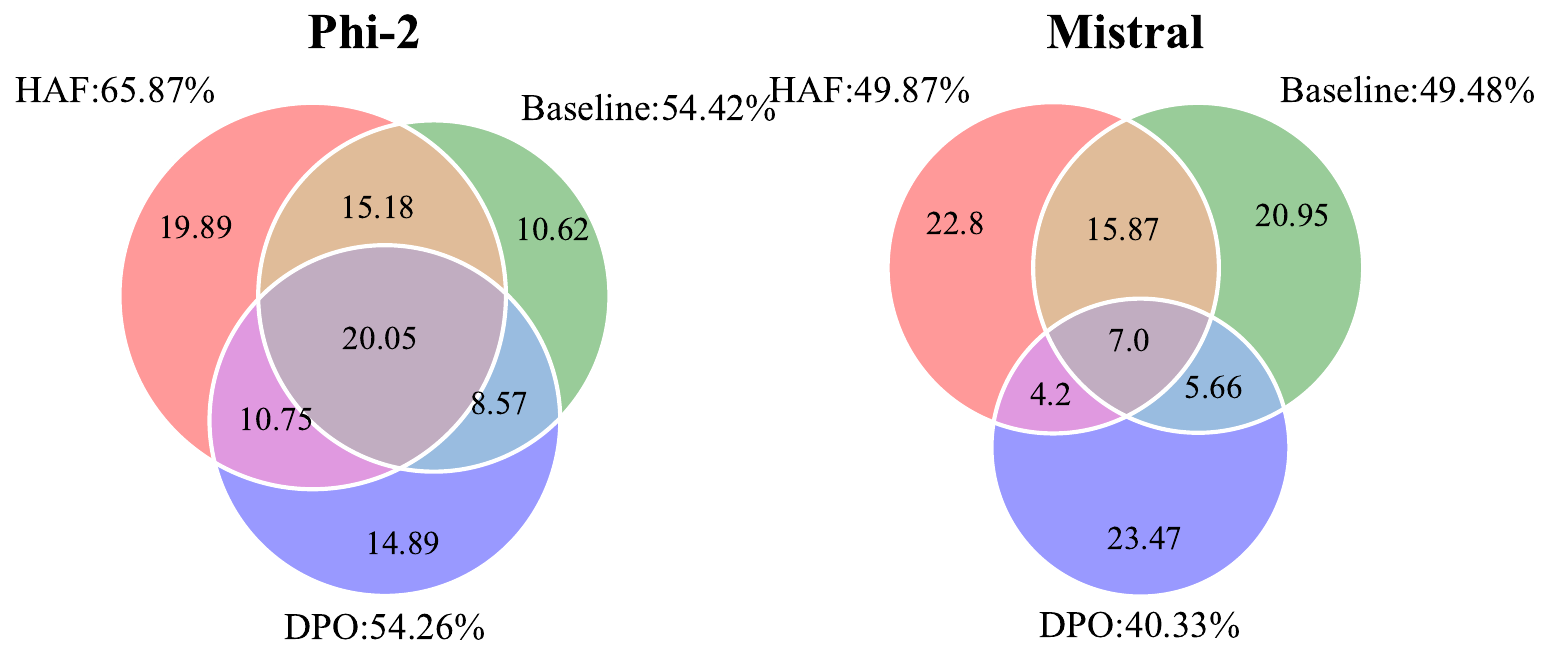}
        \caption{Avg.}
    \end{subfigure}

    \caption{Win rates on each dataset judged by \texttt{ArmoRM-Llama3-8B-v0.1}}
    \label{fig:bon_armo}
\end{figure*}

In Table~\ref{tab:total_bon} we list the recall value on each dataset. We show in Figure~\ref{fig:detailed-venn} and Figure~\ref{fig:bon_armo} the win rates on each dataset judged by \texttt{gpt-4-turbo-2024-04-09} and \texttt{ArmoRM-Llama3-8B-v0.1}~\citep{ArmoRM}, respectively.
\label{app:bon}

\section{GPT Judgement}
\label{app:gpt-4}
\paragraph{Comparing two responses} The prompt we used for judgement is listed in Table~\ref{tab:prompt for compare}. The sentence between ``<SYSTEM PROMPT>'' is the system prompt, and the others are the user prompt. ``\textbf{\{question\}}'', ``\textbf{\{response 1\}}'', ``\textbf{\{response 2\}}'' will be replaced with the actual query or responses respectively. As GPT does not exhibit a strong ``positional bias''~\citep{positionalbias}, so we just randomly interchange the order of the two responses rather than prompting twice with the responses swapped. 

\paragraph{Ranking responses} Table~\ref{tab:consumption} shows the consumption approximation for getting top-1, top-2 responses and the complete order out of 4/8 responses. We consider that performing a single sorting operation on eight responses with the model may result in a loss of precision. Besides, while binary comparisons exhibit high accuracy, repeated binary comparisons inevitably lead to cumulative errors and erroneous outcomes. Therefore, whether from a cost or accuracy standpoint, it is not a favorable option. In practice, we obtain the top 2 responses by ranking 4 responses with GPT-3.5-turbo at once. For 8 candidate responses, we first evenly divide them into two groups and use GPT to rank the responses of each group, then we rank the two sets of the top 2 responses to get the top 2 responses among 8 candidates.
\begin{figure}[h]
\centering
\includegraphics[width=0.48\textwidth]{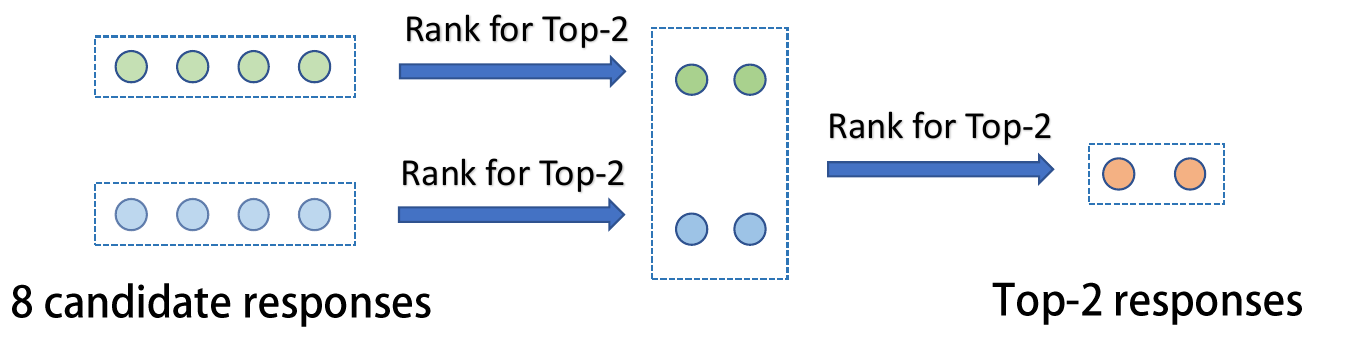}
\caption{Three times of interactions with GPT to get top-2 responses}
\end{figure}

The prompt for ranking four responses is shown in Table~\ref{tab:prompt for ranking}. GPT's answer will be parsed in JSON format.

\begin{table*}
\centering
\begin{tabular}{@{}lcccccc@{}}
\toprule
&\multicolumn{2}{c}{Top-1} & \multicolumn{2}{c}{Top-2} & \multicolumn{2}{c}{Complete sort} \\
\multicolumn{1}{r}{\# responses} & 4 & 8 & 4 & 8 & 4 & 8 \\\midrule
binary comparison & $\textbf{6}_{\,3\times2}$ & $\textbf{14}_{\,7\times2}$ & $\textbf{8}_{\,4\times2}$ & $\textbf{20}_{\,10\times2}$ & $\textbf{10}_{\,5\times2}$ & $\textbf{32}_{\,16\times2}$ \\ 
rank 4 responses & $\textbf{4}_{\,1\times4}$ & $\textbf{12}_{\,3\times4}$ & $\textbf{4}_{\,1\times4}$ & $\textbf{12}_{\,\phantom{1}3\times4}$ & $\textbf{\phantom{1}4}_{\,1\times4}$ & $\textbf{20}_{\,\phantom{1}5\times4}$ \\ 
rank 8 responses & $\textbf{4}_{\,1\times4}$ & $\textbf{\phantom{1}8}_{\,1\times8}$ & $\textbf{4}_{\,1\times4}$ & $\textbf{\phantom{1}8}_{\,\phantom{1}1\times8}$ & $\textbf{\phantom{1}4}_{\,1\times4}$ & $\textbf{\phantom{1}8}_{\,\phantom{1}1\times8}$ \\\bottomrule 
\end{tabular}
\caption{Approximation for resources consumption. The first column is three different ways of interacting with GPT. The first row is the target response(s) and the second row is the number of candidate responses. ``$a\times b$'' means we should engage with GPT-3.5 a total of $a$ times, with each interaction requiring an input of $b$ responses. For example, ``$\textbf{6}_{\,3\times2}$'' means when using binary comparison, to get the top-1 response among 4 candidate responses, we need 3 turns of interactions with each turn requiring an input of 2 responses, hence our expenditure amounts to approximately 6 units}
\label{tab:consumption}
\end{table*}

\begin{center}
\begin{table*}
    \begin{tcolorbox}
[colback=black!5!white,colframe=gray!15!gray,width=0.9\textwidth,,title={Prompt for comparing two responses.}]
\small
<SYSTEM PROMPT>You are a helpful instruction-following assistant that prints the best model by selecting the best outputs for a given instruction.<SYSTEM PROMPT>\\
Select the output (a) or (b) that best matches the given instruction. Choose your preferred output, which can be subjective. Your answer should ONLY contain: Output (a) or Output (b). \\
Here's an example:\\

 \# Example:\\
 \#\# Instruction:\\
 Give a description of the following job: "ophthalmologist"\\

 \#\# Output (a):\\
 An ophthalmologist is a medical doctor who pokes and prods at your eyes while asking you to read letters from a chart.\\

 \#\# Output (b):\\
 An ophthalmologist is a medical doctor who specializes in the diagnosis and treatment of eye diseases and conditions.\\

 \#\# Which is best, Output (a) or Output (b)?\\
 Output (b)\\

 Here the answer is Output (b) because it provides a comprehensive and accurate description of the job of an ophthalmologist. In contrast, output (a) is more of a joke.\\

 \# Task:\\
 Now is the real task, do not explain your answer, just say Output (a) or Output (b).\\

 \#\# Instruction:\\
 \{question\}\\

 \#\# Output (a):\\
 \{response 1\}\\

 \#\# Output (b):\\
 \{response 2\}\\
 
 \#\# Which is best, Output (a) or Output (b)?
 \end{tcolorbox}
 \caption{We use 1-shot for response comparison.}
 \label{tab:prompt for compare}
\end{table*}
\end{center}

\begin{center}
\begin{table*}
\tcbset{colback=black!5!white,colframe=gray!15!gray,width=0.9\textwidth,title={Prompt for ranking four responses.}}
    \begin{tcolorbox}

\small
<SYSTEM PROMPT>You are a helpful assistant, that ranks models by the quality of their answers<SYSTEM PROMPT>

I want you to create a leaderboard of different models. To do so, I will give you the instructions (prompts) given to the models, and the responses of four models. Please rank the models based on which responses would be preferred by humans. All inputs and outputs should be python dictionaries.\\

Here is the prompt:\\
\{\\
\text\qquad "instruction": \{question\},\\
\}\\

Here are the outputs of the models:\\
\text{[}\\
\text\qquad\{\\
\text\qquad\qquad"model": "model\_1",\\
\text\qquad\qquad"answer": \{output\_1\}\\
\text\qquad\},\\
\text\qquad\{\\
\text\qquad\qquad"model": "model\_2",\\
\text\qquad\qquad"answer": \{output\_2\}\\
\text\qquad\},\\
\text\qquad\{\\
\text\qquad\qquad"model": "model\_3",\\
\text\qquad\qquad"answer": \{output\_3\}\\
\text\qquad\},\\
\text\qquad\{\\
\text\qquad\qquad"model": "model\_4",\\
\text\qquad\qquad"answer": \{output\_4\}\\
\text\qquad\}\\
\text{]}\\

Now please rank the models by the quality of their answers, so that the model with rank 1 has the best output. Then return a list of the model names and ranks, i.e., produce the following output:\\
\text{[}\\
\text\qquad\{"model": "model\_1", "rank": <model-rank>\},\\
\text\qquad\{"model": "model\_2", "rank": <model-rank>\},\\
\text\qquad\{"model": "model\_3", "rank": <model-rank>\},\\
\text\qquad\{"model": "model\_4", "rank": <model-rank>\}\\
\text{]}\\

Your response must be a valid Python dictionary and should contain nothing else because we will directly execute it in Python. Please provide the ranking that the majority of humans would give.
    \end{tcolorbox}
    \caption{We rank four responses in order of quality in a single interaction.}
    \label{tab:prompt for ranking}
\end{table*}

 \end{center}

\begin{table*}
\centering
\begin{tabular}{@{}lcccccccccc@{}}
\toprule
& \multicolumn{2}{c}{AHP} & \multicolumn{2}{c}{BS} & \multicolumn{2}{c}{CA} & \multicolumn{2}{c}{Helpful} & \multicolumn{2}{c}{Harmless}\\
& Top-1 & Top-2 & Top-1 & Top-2 & Top-1 & Top-2 & Top-1 & Top-2 & Top-1 & Top-2\\
\midrule
Phi-2$_\mathrm{HAF}$ & \textbf{28.68} & \textbf{52.51}& \textbf{32.69} & \textbf{53.35}& \textbf{37.52} & \textbf{66.21}& \textbf{45.44} & \textbf{74.26}& \textbf{24.52} & \textbf{45.15}\\
Phi-2$_\mathrm{baseline}$ & 15.46 & 34.64& 29.28 & 49.72& 27.83 & 51.68& 43.62 & 73.92& 17.22 & 37.29\\
\midrule
Mistral$_\mathrm{HAF}$ & \textbf{17.42} & \textbf{31.22}& \textbf{9.94} & \textbf{17.70}& 16.00 & 28.81& \textbf{13.68} & \textbf{27.57}& \textbf{9.50} & \textbf{21.07}\\
Mistral$_\mathrm{baseline}$ & 10.97 & 23.87& 7.45 & 17.08& \textbf{17.99} & \textbf{32.78} & 12.68 & 26.36& 8.68 & 17.36\\
\bottomrule 
\end{tabular}
\caption{Top-k recall for best-of-N sampling on each dataset. The results are presented as the percentage of the chosen responses included in top-k responses. }
\label{tab:total_bon}
\end{table*}
 
\end{document}